\documentclass[10pt]{article} %
\usepackage[preprint]{tmlr}

\usepackage{amsmath,amsfonts,bm}

\def\eqref#1{equation~\ref{#1}}

\def\1{\bm{1}}

\DeclareMathAlphabet{\mathsfit}{\encodingdefault}{\sfdefault}{m}{sl}
\SetMathAlphabet{\mathsfit}{bold}{\encodingdefault}{\sfdefault}{bx}{n}

\newcommand{\E}{\mathbb{E}}

\usepackage{hyperref}
\usepackage{url}
\usepackage{amsmath,amssymb,amsthm}
\usepackage{booktabs}
\usepackage{array}
\newcolumntype{L}[1]{>{\raggedright\arraybackslash}p{#1}}
\usepackage{graphicx}
\usepackage{xcolor}
\usepackage{enumitem}
\usepackage{multirow}
\usepackage[capitalise,noabbrev]{cleveref}

\graphicspath{{figures/}}

\newtheorem{proposition}{Proposition}
\newtheorem{assumption}{Assumption}

\theoremstyle{definition}

\newcommand{\Prob}{\mathbb{P}}

\newcommand{\ind}{\mathbb{1}}

\title{Sample, Mix, or Verify? Allocating Test-Time\\ Compute under Imperfect Verifiers}

\author{\name Srinivasan Iyengar \\
      \addr Microsoft Research India}

\def\month{MM}  %
\def\year{YYYY} %
\def\openreview{\url{https://openreview.net/forum?id=XXXX}} %

\begin{document}

\maketitle

\begin{abstract}
A common way to make language models and agents more accurate is to spend more compute at
inference time: generate many candidate solutions and submit the one a verifier prefers. Success
then requires that some candidate is correct and that the verifier picks it. The first grows
predictably with the number of samples; the second does not, because real verifiers (unit tests,
majority votes, LLM judges, reward models) accept wrong answers, reject right ones, and make
correlated mistakes. With a fixed budget, a practitioner must decide how many samples to draw,
which models to draw them from, and how much to spend on verification, and today these choices are
made by rules of thumb.

We introduce \textbf{SPLIT}, which makes them from a small labelled pilot. It rests on an exact formula for
the probability of submitting a correct answer under accept/reject and top-score selection that
accounts for varying task difficulty, blind spots shared across models, and correlated repeated
verifier calls. SPLIT fits the formula by empirical Bayes and uses it to plan budgets on unseen
tasks. Across seven benchmarks spanning math, science QA, code and software-engineering agents, 26
models, 175 agent systems and six kinds of real verifiers, all from public releases, SPLIT lands
within 0.02--0.8 percentage points of the best allocation in hindsight, while every fixed rule we
test loses at least 2.5 points in some setting. Regret depends only on how well SPLIT ranks
configurations, and it stays near-optimal under compute, price and latency costs and online. The
formula also explains when asking more LLM judges lowers precision. As a forecaster far beyond the
pilot, SPLIT beats power-law fitting when the verifier caps success, and we report where it loses.
\end{abstract}

\section{Introduction}
\label{sec:intro}

\begin{figure}[t]
  \centering
  \includegraphics[width=\linewidth]{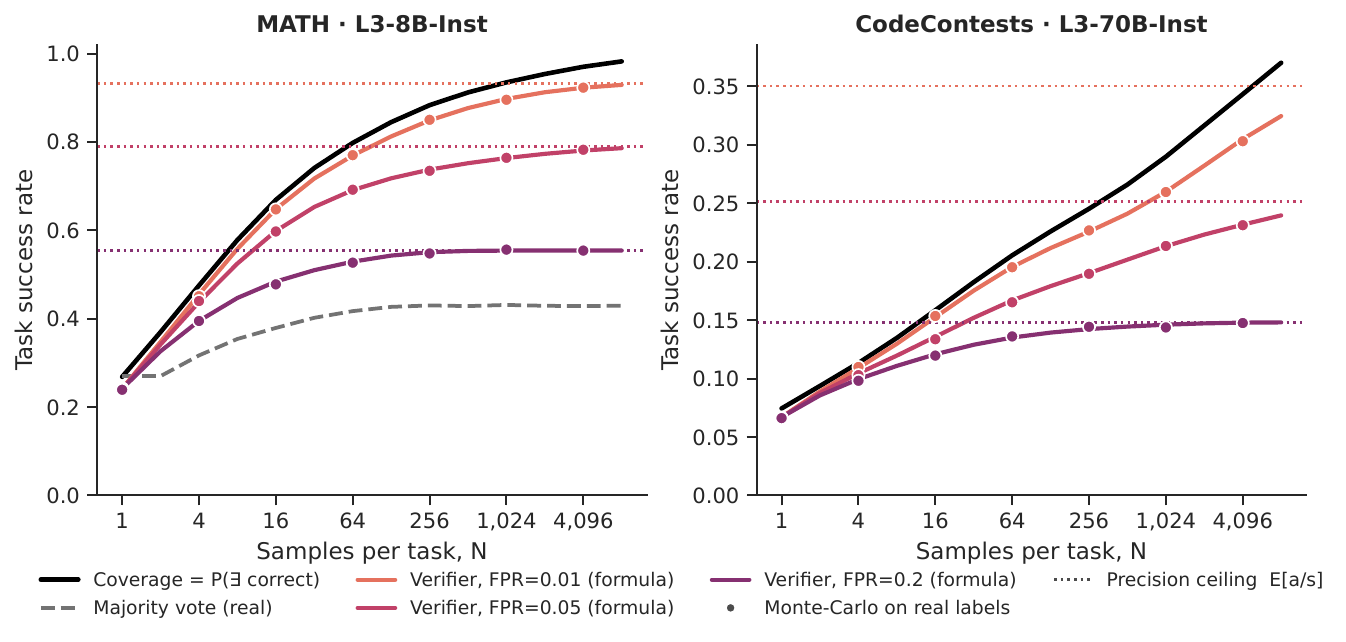}
  \caption{\textbf{Generation is not selection.} Coverage (black) keeps rising with the number of
  samples $N$, but verified success flattens at a ceiling set by the verifier's precision
  (dotted). Coloured lines are the exact success formula (\cref{prop:law}); markers are simulations of the
  verifier on the \emph{real} correctness labels of the Large Language Monkeys release
  \citep{brown2024monkeys}. A real majority vote (grey dashed) stalls near 43\% on MATH while
  coverage reaches 98\%.}
  \label{fig:gen-vs-sel}
\end{figure}

Spending more compute at inference time, by sampling many candidate solutions and keeping one, is
now a standard way to raise the accuracy of language models and agents
\citep{cobbe2021verifiers,wang2023selfconsistency,brown2024monkeys,snell2024scaling}. Coverage, the
probability that at least one of $N$ attempts is correct, grows smoothly with $N$, often as a
power law over several orders of magnitude \citep{brown2024monkeys,schaeffer2025monkeys}. But a
system is graded on the one answer it submits. That answer is chosen by a \emph{verifier}: unit
tests, a majority vote, an LLM judge or a reward model. Real verifiers accept some wrong answers
and reject some right ones, and their mistakes are not independent. The same plausible wrong
answer tends to fool every judge, and different generators tend to fail on the same hard problems.
As a result, verified success can stall far below coverage. On MATH, majority voting over
Llama-3-8B-Instruct samples stays near 43\% while coverage reaches 98\% (\cref{fig:gen-vs-sel}), and
\citet{stroebl2024limits} show that imperfect unit tests cap what resampling can achieve.

Practitioners therefore face an allocation problem rather than a scaling problem. With a fixed
budget, should they draw more samples, mix in a second model, spend calls on verification, or
change how strictly the verifier accepts? Fixed rules of thumb answer these questions
differently: ``use the strongest model'', ``scale solutions faster than verifications''
\citep{singhi2025solve}, ``ask more judges''. We show that each of them is badly wrong in some
realistic setting. We argue that the decision should be made from a small labelled \emph{pilot}
using a model of \emph{selection}, not only of generation.

\paragraph{Three obstacles.} Fixed rules fail for three reasons, and our method addresses each.
(1)~\emph{The verifier caps what is handed in.} When a verifier sometimes accepts wrong answers,
verified success saturates at the verifier's precision however many samples are drawn
(\cref{sec:law}), so the verifier's true- and false-accept rates must be modelled, not only how
often the generator is right. (2)~\emph{Tries are correlated.} Llama-3-8B-Instruct solves an average
MATH problem 27\% of the time, which would give 96\% coverage at 10 tries if every problem were
equally hard; its measured per-problem solve rates give 61\%, because 44 of the 128 problems are
almost never solved. A second model gets stuck on the same problems (per-problem solve rates of
Llama-3-8B-Instruct and Gemma-7B correlate at 0.80), and a third of wrong MATH-500 answers fool at
least six of eight LLM judges, so asking again repeats the mistake. (3)~\emph{No fixed split works
everywhere.} The best configuration changes with the budget, the models and the verifier
(\cref{sec:res-alloc}), so it has to be chosen per setting.

\paragraph{Approach.} We write success as
$\Prob(\text{success}) = \Prob(\text{a correct candidate exists}) \times \Prob(\text{it is selected}
\mid \text{it exists})$ and model the second factor explicitly. \Cref{sec:law} derives an exact
success formula for two selectors used in practice, uniform choice among accepted candidates and
choice of the top-scoring candidate. The formula models the verifier's error rates (obstacle 1) and the
three sources of correlation (obstacle 2): heterogeneous task difficulty, shared blind spots across
generators (a Gaussian copula), and correlated repeated verifier calls (a candidate-level acceptance
propensity).
\Cref{sec:estimation} estimates the formula's ingredients from a pilot by empirical Bayes, including
variants for verifiers whose accuracy depends on the task. A planner then chooses the
configuration that maximises predicted success under a cost budget (obstacle 3), and is evaluated on tasks
not used for calibration. We call the whole method \textbf{SPLIT}, because it chooses how to split a fixed budget across
generators, samples and verification.

\paragraph{Contributions.}
\begin{enumerate}[leftmargin=*,itemsep=2pt,topsep=2pt]
  \item \textbf{SPLIT, a pilot-calibrated allocator of test-time compute.} One planner chooses
  generators, sample counts, verifier calls or verifier thresholds under a budget. Across five
  allocation problems with real or simulated verifiers, plus an agentic threshold-selection problem,
  it lands within 0.02--0.8 points of the hindsight-optimal allocation on unseen tasks. Every fixed
  rule we compare against, including our adaptation of the compute-optimal exponents of
  \citet{singhi2025solve}, loses at
  least 2.5 points somewhere (\cref{sec:res-alloc}). Its regret is bounded by differential rather
  than absolute forecast error (\cref{prop:regret}), it stays within 0.62 points under five ways of
  pricing a sample, and an online variant that re-plans per task matches the best static
  configuration in hindsight (\cref{sec:res-beyond}).
  \item \textbf{An exact success formula} with heterogeneous difficulty, cross-generator
  correlation and correlated verifier calls, for both accept/reject and argmax selection, with
  proofs (\cref{sec:law}, \cref{app:proofs}).
  \item \textbf{Three mechanisms, measured on real verifiers} (\cref{sec:res-mech}).
  (i)~We give the exact condition under which majority-of-$k$ re-verification \emph{lowers}
  attainable precision, and show that real LLM judges meet it on MATH-500 but not on GPQA.
  (ii)~Pooling verifier accuracy across tasks biases forecasts because judge accuracy is coupled to
  task difficulty; modelling this coupling recovers the measured gain from extra judge calls.
  (iii)~We replicate the rise of verifier false-positive rates with generator capability
  \citep{stroebl2024limits} on HumanEval+, MBPP+ and CodeContests, and show that for scored
  verifiers it is a threshold-calibration effect.
  \item \textbf{A candid comparison with curve fitting} (\cref{sec:res-forecast}). Using SPLIT
  as a point forecaster beats power-law extrapolation clearly under a simulated verifier and on
  agentic SWE-Bench Verified, modestly but significantly on code with real tests, and not at all
  for pure coverage or for argmax selection. This includes a long-horizon test on a pool of 175
  SWE-bench agent systems.
\end{enumerate}

All experiments use publicly released samples and verdicts. We spend no API budget, re-execute
80{,}403 CodeContests programs locally, and report paired significance tests for every headline
comparison (\cref{app:significance}). The code, which reproduces every number and figure, will be
released publicly.

\section{Related work}
\label{sec:related}

\paragraph{Repeated sampling and coverage laws.} \citet{brown2024monkeys} show that coverage grows
roughly as a power law in the number of samples across math, code and formal proofs, and document
that majority voting and reward-model selection plateau well below coverage.
\citet{schaeffer2025monkeys} explain the power law through a heavy-tailed distribution of
per-problem success rates. Both works model \emph{coverage}. We add a predictive, mechanistic model for
the selection factor, a treatment of correlated errors, and pilot-calibrated allocation; our
coverage component uses the same beta-binomial view of difficulty. \citet{chen2024morecalls} show that voting accuracy can be
non-monotone in the number of calls, depending on the mix of easy and hard queries. We obtain a
related non-monotonicity for \emph{verification} calls, with an exact condition
(\cref{prop:reverify}).

\paragraph{Imperfect verifiers.} \citet{stroebl2024limits} show that false positives of unit tests
bound the accuracy of resampling, that a model's single-sample accuracy is strongly correlated with
its false-positive rate on HumanEval and MBPP, and that the best number of attempts is often small.
We build on their observations. We replicate the accuracy--FPR correlation on HumanEval+, MBPP+ and
CodeContests and refine it for scored verifiers (\cref{sec:res-mech}). We also turn the bound into
a predictive model with correlated errors, estimated from a pilot, and use it for allocation.
Verifier design is a large literature: outcome and process reward models
\citep{cobbe2021verifiers,lightman2023verify}, generative verifiers \citep{zhang2024genrm}, and
weak-verifier ensembles such as Weaver \citep{saadfalcon2025weaver}. That work builds better
verifiers. We take a verifier as given, measure its errors, and decide how much to rely on it.

\paragraph{Compute-optimal test-time scaling.} \citet{snell2024scaling} allocate test-time compute
adaptively between search and revision based on estimated question difficulty.
\citet{singhi2025solve} fit compute-optimal power laws for the split between solutions and
generative verifications ($S \propto C^{0.57}$, $V \propto C^{0.39}$). We use their rule as a
baseline. Our approach differs in three ways. It is mechanistic rather than a curve fit, so it
transfers to new verifiers and budgets. It models cross-generator correlation and the correlation
of repeated verifier calls. And it chooses jointly over generators, samples, verifier calls and
thresholds.

\paragraph{Agents and best-of-$N$ for software engineering.} SWE-bench \citep{jimenez2024swebench},
SWE-bench Verified \citep{swebenchverified2024} and agent scaffolds such as SWE-agent
\citep{yang2024sweagent} made repository-level evaluation standard. R2E-Gym \citep{jain2025r2egym}
and CodeMonkeys \citep{ehrlich2025codemonkeys} scale the number of rollouts and select among them
with execution-free verifiers or generated tests. Holistic agent evaluation \citep{kapoor2025hal}
and studies of multi-agent scaling \citep{kim2025scalingagents,yang2026diversity} show that
outcomes depend on the scaffold and on the diversity of agents, not only on their number. We use
the public rollouts of R2E-Gym and the per-instance results of 175 SWE-bench submissions
\citep{swebenchexperiments} to test SPLIT on agents.

\paragraph{Statistical tools.} The estimators are standard empirical Bayes
\citep{efron2010largescale}, with beta-binomial and non-parametric maximum-likelihood
\citep{kiefer1956npmle} priors for difficulty. The diversity result uses the supermodular order of
Gaussian copulas \citep{muller2000supermodular}, and the re-verification result is a version of
the Condorcet jury theorem with correlated votes \citep{ladha1992condorcet}.

\section{The success formula for verified best-of-\texorpdfstring{$N$}{N}}
\label{sec:law}

\paragraph{Setting.} A task $t$ is attempted by generators $m=1,\dots,M$; a sample from $m$ is correct
with probability $p_{t,m}$. A verifier accepts a correct candidate with probability $\tau$ (true
positive rate, TPR) and a wrong one with probability $f_t$ (false positive rate, FPR). Both may
depend on the task. Write $a = \tau p$ and $s = a + f(1-p)$ for the per-sample probabilities of
``accepted and correct'' and ``accepted''. The \emph{uniform-among-accepted} selector submits a
uniformly random accepted candidate and abstains (counted as failure) if none is accepted. Coverage
is $C_N = 1-(1-p)^N$. Expectations $\E$ are over the task distribution. Throughout we assume:

\begin{assumption}[Conditional independence]\label{ass:ci}
Given the task, candidates are independent: candidate $i$ from generator $m$ has a
(correct, accepted) pair $(C_i, A_i)$ with $\Prob(C_i{=}1, A_i{=}1)=a_m$ and $\Prob(A_i{=}1)=s_m$,
independently across candidates. Sample counts $n_m$ are non-negative integers; with no samples,
or with $s=0$, the selector abstains.
\end{assumption}

All dependence \emph{across tasks} (difficulty, cross-generator correlation, task-dependent verifier
rates) enters through the joint distribution of $(p_{t,1},\dots,p_{t,M}, \tau_t, f_t)$ over tasks.
Dependence \emph{within a candidate} across repeated verifier calls enters through a
candidate-level propensity (\cref{prop:reverify}).

\begin{proposition}[Exact success formula]\label{prop:law}
Under \cref{ass:ci}, with $n_m$ samples from generator $m$, the probability that the selected
candidate is correct is
\begin{equation}\label{eq:law}
  S \;=\; \int_0^1 \sum_{m} n_m\, a_m\, (1-s_m u)^{n_m-1} \prod_{l\ne m} (1-s_l u)^{n_l}\, du .
\end{equation}
For a single generator with $s>0$, $S_N = \frac{a}{s}\bigl(1-(1-s)^N\bigr)$, which is
non-decreasing in $N$ and converges to the \emph{precision ceiling} $a/s$.
\end{proposition}

The single-generator form factorises success as (probability that something is accepted) $\times$
(precision of acceptance). The proof (\cref{app:proofs}) represents ``uniform among accepted'' as
``first accepted under independent uniform arrival times''. Averaging \cref{eq:law} over the task
distribution gives the expected success of a configuration. Two regimes follow
(\cref{prop:regimes}, \cref{app:proofs}). If the FPR is bounded below by $f_0>0$ on every task,
verified success saturates exponentially, $0 \le S_\infty - S_N \le S_\infty (1-\min(\tau,f_0))^N$.
If instead the FPR has mass near zero, $\Prob(f_t \le x) \le K x^{a_f}$, the saturation gap is
bounded only by $O(N^{-a_f})$. Coverage itself decays polynomially, $1-\E[C_N] \sim c\,\Gamma(\alpha)
N^{-\alpha}$, when the density of $p$ behaves like $c\,p^{\alpha-1}$ near zero. Finally, verified
success equals TPR-thinned coverage up to $N\,\E[f_t]$:
$0 \le \E[1-(1-\tau p)^N] - S_N \le N\,\E[f_t]$, so small budgets behave as if the verifier were
perfect and large budgets hit the ceiling.

\paragraph{Shared blind spots.} With two generators and a budget split $(1-\lambda)N$ and $\lambda N$,
the coverage of the mixture is concave in $\lambda$. Mixing in B strictly helps iff
$\E[w_A \log(1-p_B)] < \E[w_A \log(1-p_A)]$ with $w_A = (1-p_A)^N$, i.e.\ iff B is better than A
\emph{on the tasks A still fails}. If $(p_A,p_B)$ follow a Gaussian copula with correlation $\rho$
and fixed marginals, the gain from an even mix is non-increasing in $\rho$
(\cref{prop:diversity}, \cref{app:proofs}). The planner therefore needs one number per generator
pair, $\rho$, estimated from the pilot.

\paragraph{Correlated re-verification.} Repeated verifier calls on the same candidate are not
independent. A judge that is fooled by an answer once tends to be fooled again, and so do other
judges. We model each candidate $c$ by a propensity $q_c$: given $q_c$, calls are i.i.d.\
Bernoulli$(q_c)$, with $q \sim D_c$ for correct and $q \sim D_w$ for wrong candidates. The
majority of $k$ (odd) calls accepts with probability $T_k = \E_{D_c}[B_k(q)]$ for correct and
$F_k = \E_{D_w}[B_k(q)]$ for wrong candidates, where $B_k(q) = \Prob(\mathrm{Bin}(k,q) > k/2)$.
In experiments we take $D_c, D_w$ to be Beta distributions with means $\tau, f$ and a concentration
$\kappa$ ($\kappa\to\infty$: independent calls; $\kappa\to 0$: every call repeats the first).

\begin{proposition}[When re-verification helps or hurts]\label{prop:reverify}
(a)~As $k\to\infty$ over odd integers, $T_k \to T_\infty = \Prob_{D_c}(q>\tfrac12) + \tfrac12
\Prob_{D_c}(q=\tfrac12)$, and likewise $F_k \to F_\infty$. (b)~If every $q\in\{0,1\}$, then
$T_k = T_1$ and $F_k = F_1$ for all $k$. (c)~Fix a task with $0<p<1$ and positive rates, and take
$N\to\infty$. For any odd $j,k$, the precision ceiling $pT_k/(pT_k+(1-p)F_k)$ with $k$ calls is
strictly above the ceiling with $j$ calls iff $F_k/T_k < F_j/T_j$. In particular, $k$ calls
\emph{lower} the ceiling relative to one call iff
\begin{equation}\label{eq:hurt}
  \frac{F_k}{F_1} \;>\; \frac{T_k}{T_1},
\end{equation}
that is, iff majority voting inflates false acceptances by a larger factor than true acceptances.
The same holds for $k=\infty$ with the limits of part (a). A larger FPR ($F_k > F_1$) alone is not
sufficient.
\end{proposition}

Majority voting amplifies whatever most calls believe. If most wrong answers fool most judges
($D_w$ has mass above $\tfrac12$), $F_k$ grows with $k$. Whether this hurts depends on how much the
same amplification helps correct answers. The condition concerns the large-$N$ precision ceiling;
at finite $N$, extra calls also change how often nothing is accepted. \Cref{app:proofs} gives an
example where the ceiling falls and a counterexample with $F_\infty > F_1$ where it rises.

\paragraph{Argmax selection.} Many systems instead submit the candidate with the highest verifier
score. Let correct and wrong candidates have score distributions $G_c$ and $G_w$, and
$H = pG_c + (1-p)G_w$.

\begin{proposition}[Verified success under argmax selection]\label{prop:argmax}
For $N\ge 1$ i.i.d.\ candidates, (a)~if $H$ is atomless and $G_c$ has density $g_c$,
$P_N = Np\,\E_{Y\sim G_c}[H(Y)^{N-1}]$. (b)~With scores binned into ordered bins, let $r_c[b]$ and
$r_w[b]$ be the \emph{joint} probabilities that a candidate is correct and in bin $b$, or wrong and
in bin $b$ (so $\sum_b (r_c[b]+r_w[b]) = 1$), $r = r_c + r_w$, $R_b = \sum_{b'\le b} r[b']$ and
$R_{-1}=0$. With ties broken uniformly, $P_N = \sum_b (R_b^N - R_{b-1}^N)\, r_c[b]/r[b]$, where
empty bins contribute zero. (c)~$P_N$ converges to the precision of
the top occupied bin. In the continuous case it converges to $L$ whenever the local precision
$\eta(y) = p g_c(y) / (p g_c(y) + (1-p) g_w(y))$ has a limit $L$ at the upper end of the score
support.
\end{proposition}

So argmax selection replaces the \emph{average} precision $a/s$ among accepted candidates with the
\emph{local} precision at the top of the score distribution. Which one is larger is an empirical
question. It depends on the upper tails of $G_c$ and $G_w$.

\section{From a pilot to a plan}
\label{sec:estimation}

The success formula needs the joint distribution of difficulty and verifier behaviour over tasks. We estimate
it from a labelled pilot: $n$ samples per task (typically 8--64) on a set of calibration tasks,
with correctness labels and verifier verdicts or scores. All estimators are empirical Bayes; full
details are in \cref{app:estimation}.

\paragraph{Difficulty.} For each generator, pilot success counts $k_t \sim
\mathrm{BetaBin}(n,\alpha,\beta)$ give a beta prior on $p_{t}$ by maximum likelihood, optionally
zero-inflated for tasks no sample ever solves. For long horizons we also use the
non-parametric maximum-likelihood estimator (NPMLE) of the difficulty distribution
\citep{kiefer1956npmle}, which can put point masses at $p=0$. Per-task posteriors combine the prior
with the task's own pilot counts.

\paragraph{Cross-generator correlation.} For each pair of generators we fit the Gaussian-copula
correlation $\rho$ by maximising the bivariate likelihood of the pair's pilot counts, integrating
the latent normals by Gauss--Hermite quadrature. We then project the correlation matrix onto the
nearest valid correlation matrix and draw correlated $(p_{t,1},\dots,p_{t,M})$ by Monte Carlo.

\paragraph{Verifier.} The pooled estimator uses a single TPR and an empirical-Bayes beta prior over
per-task FPRs. For repeated stochastic verification (LLM judges) we fit the propensity
concentration $\kappa$ separately for correct and wrong candidates from the spread of acceptance
counts across repeated calls. Three refinements address verifiers whose accuracy depends on the
task:
\begin{itemize}[leftmargin=*,itemsep=1pt,topsep=1pt]
  \item \textbf{Task-varying verifier.} Each task has its own $(p_t,\tau_t,f_t)$, with beta
  marginals coupled by a three-dimensional Gaussian copula fitted on the pilot. This captures, for
  example, LLM patch verifiers that are lenient on easy issues.
  \item \textbf{Automatic model choice.} Neither the pooled nor the task-varying verifier model is
  best everywhere. We fit both on half of each task's pilot samples, predict verified success at the
  full pilot size, and keep the model whose prediction is closer to the model-free success on the
  full pilot. In-sample BIC chose the wrong model on all three datasets where we tried it, so we do
  not use it.
  \item \textbf{Empirical task population (for judges).} Instead of a parametric difficulty prior
  combined with pooled judge rates, SPLIT averages over the calibration tasks' own pilot
  candidates and their measured judge propensities. This keeps the coupling between task
  difficulty and judge accuracy, which pooling destroys (\cref{sec:res-mech}). A judge-level item
  response model (each judge's leniency, each candidate's deceptiveness, each task's effect) is
  reported as an alternative.
\end{itemize}

\paragraph{Argmax selection.} We bin verifier scores and estimate the class-conditional score
distributions from the pilot, then evaluate \cref{prop:argmax}(b). Pooling scores across tasks
overstates how well the verifier separates good from bad candidates \emph{within} a task: on
SWE-Bench Verified a task's mean score correlates with its solve rate ($r=0.59$), and within-task
AUC is 0.75 against 0.79 pooled. We therefore fit a per-task random intercept on logit scores by EM
and keep the empirical residual tails.

\paragraph{Planner.} A configuration specifies generators with mixing weights $w_m$, the number of
verifier calls $k$ (or an acceptance threshold), and implicitly the sample counts
$n_m = \lfloor w_m B / (c_m + k c_v) \rfloor$ for a per-task budget $B$, generation costs $c_m$ and
verifier cost $c_v$. The planner evaluates \cref{eq:law} (or \cref{prop:argmax}) under the fitted
posterior for every configuration in a grid and returns the one with the highest predicted
success. Costs are either a parameter-count proxy (an 8B-parameter sample costs 1) or
\emph{compute-matched}: parameters $\times$ measured mean output tokens for the model on that
dataset, normalised to Llama-3-8B-Instruct. To cover money and time as well as compute, we also
price a sample as $(\text{parameters})^{\gamma}\times$ output tokens: $\gamma=1$ is compute,
$\gamma\approx0.7$ mimics per-token prices that grow sublinearly with model size, $\gamma\approx0.5$
mimics decode latency when larger models are served on more accelerators, and $\gamma=0$ is a flat
per-token price. Judge calls cost one generation, test execution costs nothing, and a simulated
verifier call costs 0.25.

\paragraph{What accuracy the planner needs.} The planner only has to \emph{rank} configurations, so
an error shared by all configurations is harmless. Let $\mathcal C$ be the configurations at a
budget, $S(c)$ the true success on the evaluation tasks, $\hat S(c)$ the prediction,
$e(c)=\hat S(c)-S(c)$, $\hat c=\arg\max_c\hat S(c)$, $c^\star=\arg\max_c S(c)$, and
$\operatorname{osc}(e)=\max_c e(c)-\min_c e(c)$.

\begin{proposition}[Planner regret]\label{prop:regret}
\begin{enumerate}[label=(\alph*),leftmargin=*,itemsep=1pt]
  \item $\mathrm{Reg}:=S(c^\star)-S(\hat c)\le e(\hat c)-e(c^\star)\le\operatorname{osc}(e)\le2\max_c|e(c)|$.
  \item If $S(c^\star)-\max_{c\ne c^\star}S(c)>\operatorname{osc}(e)$, then $\hat c=c^\star$.
  \item Let $S_{\mathrm{cal}}(c)$ be the true success on the $T_{\mathrm{cal}}$ calibration tasks
  and $b=\hat S-S_{\mathrm{cal}}$ the model's error there. If calibration and the $T_{\mathrm{ev}}$
  evaluation tasks are i.i.d.\ draws from one population, then with probability at least $1-\delta$,
  \[
  \mathrm{Reg}\le\operatorname{osc}(b)+2\sqrt{\tfrac12\left(\tfrac1{T_{\mathrm{cal}}}+\tfrac1{T_{\mathrm{ev}}}\right)\log\tfrac{2|\mathcal C|}{\delta}}.
  \]
\end{enumerate}
\end{proposition}

Part~(a) explains why regret can be far below forecasting error, and part~(c) why it grows only
logarithmically with the size of the search space. The constant in~(c) is loose for our task
counts (see \cref{sec:res-beyond}).

\section{Experimental setup}
\label{sec:setup}

\begin{table}[t]
\centering
\caption{\textbf{Data.} All samples and verdicts come from public releases; we generate nothing
new. Benchmarks: MATH \citep{hendrycks2021math}, CodeContests \citep{li2022alphacode}, HumanEval
\citep{chen2021humaneval} with HumanEval+ \citep{liu2023evalplus}, MBPP \citep{austin2021mbpp},
MATH-500, GPQA \citep{rein2023gpqa} and SWE-Bench Verified \citep{jimenez2024swebench}. Releases
are cited in the text.}
\label{tab:data}
\small
\setlength{\tabcolsep}{4pt}
\begin{tabular}{@{}L{3.0cm}rL{1.6cm}L{3.6cm}L{4.6cm}@{}}
\toprule
Benchmark & Tasks & Samples per task & Generators & Verifier(s) \\
\midrule
MATH & 128 & 10{,}000 & 5 (Llama-3 8B/70B, Gemma 2B/7B) & simulated (TPR, FPR, $\kappa$ controlled); real majority vote \\
CodeContests & 140 & 10{,}000 & 3 (Llama-3 8B/70B-Inst., Gemma-7B) & simulated; \textbf{public tests} (80{,}403 programs executed) \\
HumanEval & 164 & 50--1{,}000 & 17 (CodeGen-6B to GPT-4o) & \textbf{base tests}; truth: HumanEval+ \\
MBPP & 399 & 20--50 & 10 & \textbf{base tests}; truth: MBPP+ \\
MATH-500, GPQA & 500, 646 & 100 & Llama-3.1-70B-Inst.\ (+8B) & \textbf{8 LLM judges} (random judge per call, majority of $k$); 19 reward models; Weaver score \\
SWE-Bench Verified (R2E-Gym rollouts) & 492 & 22 & R2E-Gym 32B agent & \textbf{LLM patch verifier} $P(\text{yes})$: thresholds or argmax \\
SWE-Bench Verified (agent pool) & 500 & 1 per system & 175 submitted systems & none (coverage only) \\
\bottomrule
\end{tabular}
\end{table}

\paragraph{Data.} \Cref{tab:data} lists the benchmarks. MATH and CodeContests samples come from the
Large Language Monkeys release \citep{brown2024monkeys}. For CodeContests we executed every unique
program against the public tests (the verifier) and against all tests (the ground truth) in
isolated subprocesses. HumanEval and MBPP samples and verdicts come from the release of
\citet{stroebl2024limits}; the original tests are the verifier and the extended EvalPlus tests the
ground truth. MATH-500 and GPQA candidates with verdicts from eight LLM judges and 19 reward-model
scores come from the Weaver release \citep{saadfalcon2025weaver}. Calling ``a random judge'' gives a
stochastic verifier whose per-candidate propensity is the fraction of judges that accept.
SWE-Bench Verified rollouts with an execution-free verifier probability come from R2E-Gym
\citep{jain2025r2egym}. We keep the 492 issues with at least 22 rollouts and draw 22 at random per
issue. The agent pool uses official per-instance resolved labels of the 175 systems in the
SWE-bench experiments repository \citep{swebenchexperiments}.

\paragraph{Protocols.}
\begin{itemize}[leftmargin=*,itemsep=1pt,topsep=1pt]
  \item \textbf{Allocation on unseen tasks.} Calibrate on a random 50\% of tasks from an $n$-sample
  pilot (32 unless stated), plan for each budget, and score the chosen configuration on the other
  50\% against ground truth. The oracle is the best configuration in hindsight on the evaluation
  tasks; regret is oracle success minus achieved success, in percentage points. We use 5 splits
  (simulated verifier) or 10 (real verifiers) and budgets of 4--256 (sampling) or 16--1{,}024 (judges)
  cost units per task. The search space covers single generators and pairs with mixing weights
  $\{\tfrac14,\tfrac12,\tfrac34\}$, $k\in\{1,3,5,9\}$ verifier calls, and, for SWE-bench, thresholds
  $\{$none$, 0.3, 0.5, 0.7\}$ for $N\in\{1,2,4,8,16\}$.
  \item \textbf{Pilot extrapolation.} Draw $n \in \{8,16,32,64\}$ pilot samples per task
  (\{2,4,6\} rollouts on SWE-bench; 4--32 systems for the agent pool) and predict success for all
  $N$ up to the held-out horizon: 8{,}192 for MATH and CodeContests, about 128 for HumanEval, 16
  for SWE-bench rollouts and up to 167 for the agent pool. Truth is computed on held-out samples
  (exact hypergeometric pass@$N$ for coverage). The metric is the mean absolute error (MAE) over
  $N$ beyond the pilot, over 6--10 random pilots.
  \item \textbf{Statistics.} Headline comparisons use paired Wilcoxon signed-rank tests and
  bootstrap 95\% intervals over shared (model, pilot, seed) or (seed, budget) units
  (\cref{app:significance}).
\end{itemize}

\paragraph{Baselines.} For allocation: \emph{strongest model}, \emph{most accurate per cost},
\emph{uniform ensemble}, a heuristic adaptation of the compute-optimal exponents of
\citet{singhi2025solve} ($S\propto C^{0.57}$, $V\propto C^{0.39}$, anchored at $V=1$ at the smallest
budget, with the best pilot generator; their study concerns the solve-versus-verify split for one
generator with generative verification, so conclusions about this baseline concern transfer to our
settings, not their original claim),
\emph{generation only} ($k=1$), \emph{maximum verification} ($k=9$), a \emph{fixed threshold} of 0.5,
\emph{no verifier}, and SPLIT with errors assumed \emph{independent} ($\rho=0$, $\kappa=\infty$) as
an ablation. For forecasting: the power law of \citet{brown2024monkeys} fitted to the pilot's
selection curve, the same power law fitted to coverage (implicitly assuming a perfect verifier),
per-task plug-in estimates without shrinkage, and the independence formula $1-(1-\bar p)^N$.

\section{Results}
\label{sec:results}

\subsection{The pilot-calibrated planner allocates near-optimally}
\label{sec:res-alloc}

\begin{table}[t]
\centering
\caption{\textbf{Allocation regret on unseen tasks: samples and generators} (percentage points
below the hindsight-optimal configuration; mean over budgets and splits; lower is better).
``Compute-matched'' prices samples by parameters $\times$ measured output tokens instead of
parameters alone. SPLIT with independent errors is our ablation; the others are fixed rules.}
\label{tab:regret-sampling}
\small
\setlength{\tabcolsep}{5pt}
\begin{tabular}{@{}lcccccc@{}}
\toprule
& \multicolumn{2}{c}{Simulated verifier} & \multicolumn{2}{c}{Sim., compute-matched} & \multicolumn{2}{c}{CodeContests, real tests} \\
\cmidrule(lr){2-3}\cmidrule(lr){4-5}\cmidrule(lr){6-7}
Method & MATH & CodeContests & MATH & CodeContests & param.\ cost & compute-matched \\
\midrule
\textbf{SPLIT (ours)} & \textbf{0.54} & 0.82 & \textbf{0.47} & \textbf{0.65} & \textbf{0.62} & \textbf{0.46} \\
SPLIT, independent errors & 1.69 & 1.01 & 1.36 & 0.82 & 1.46 & 1.34 \\
\citet{singhi2025solve} rule & 11.95 & \textbf{0.78} & 11.94 & 0.73 & --- & --- \\
Strongest model & 12.35 & 0.99 & 12.72 & 1.02 & 0.94 & 0.94 \\
Uniform ensemble & 16.73 & 3.62 & 16.74 & 3.82 & 2.33 & 2.53 \\
Most accurate per cost & 19.81 & 5.31 & 20.67 & 4.58 & 3.74 & 3.35 \\
\bottomrule
\end{tabular}
\end{table}

\begin{table}[t]
\centering
\caption{\textbf{Allocation regret on unseen tasks: how much to verify} (percentage points; mean
over budgets/$N$ and 10 splits). Left: solve-vs-verify with real LLM judges (each judge call costs
one generation). Right: choosing the acceptance threshold of an LLM patch verifier for each
number of agent rollouts on SWE-Bench Verified.}
\label{tab:regret-verify}
\small
\setlength{\tabcolsep}{5pt}
\begin{tabular}{@{}lcc@{\hspace{1.2em}}lc@{}}
\toprule
\multicolumn{3}{c}{LLM judges (Weaver release)} & \multicolumn{2}{c}{SWE-Bench Verified agents} \\
\cmidrule(r){1-3}\cmidrule(l){4-5}
Method & MATH-500 & GPQA & Method & Regret \\
\midrule
\textbf{SPLIT, empirical task population} & 0.06 & \textbf{0.02} & \textbf{SPLIT, task-varying verifier} & \textbf{0.33} \\
SPLIT, fitted judge correlation & 0.06 & 0.11 & SPLIT, auto-selected verifier model & 0.56 \\
SPLIT, judge-level IRT & \textbf{0.04} & 0.05 & SPLIT, pooled verifier & 2.36 \\
SPLIT, independent judge calls & 0.23 & 0.53 & Fixed threshold 0.5 & 2.62 \\
\citet{singhi2025solve} rule & 0.05 & 1.20 & No verifier (random rollout) & 3.58 \\
Generation only ($k=1$) & 0.10 & 2.58 & & \\
Maximum verification ($k=9$) & 1.69 & 3.84 & & \\
\bottomrule
\end{tabular}
\end{table}

\begin{figure}[t]
  \centering
  \includegraphics[width=\linewidth]{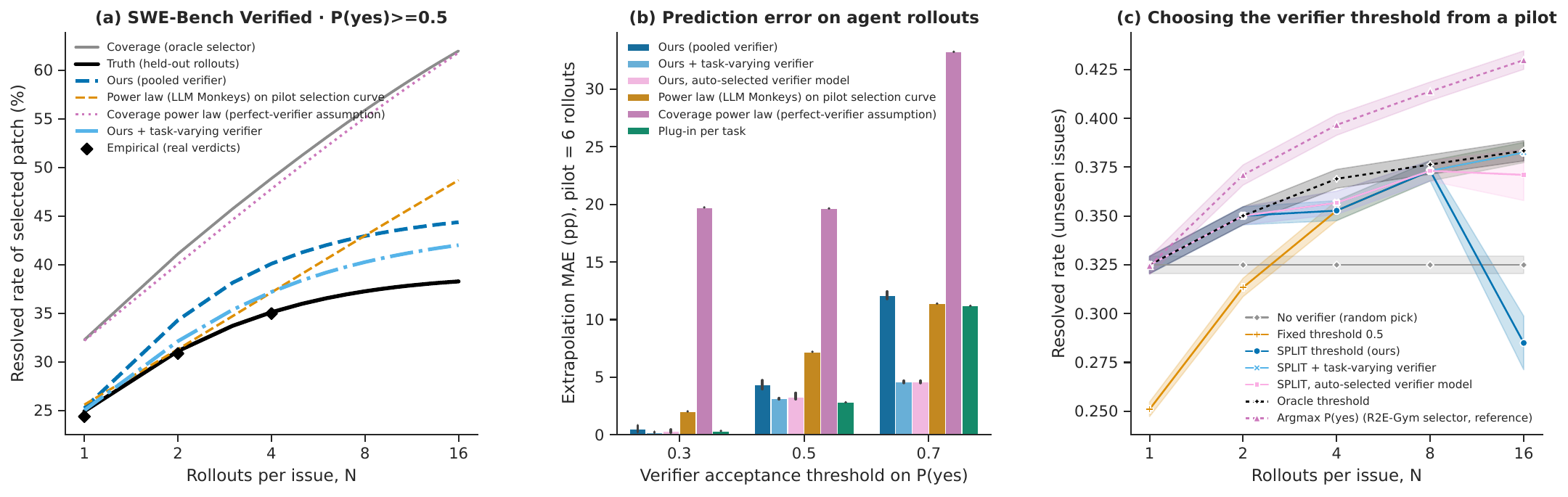}
  \caption{\textbf{Agentic setting: SWE-Bench Verified rollouts with an execution-free LLM verifier}
  (R2E-Gym release). (a)~Resolved rate of the selected patch at threshold 0.5: truth, SPLIT (pooled
  and task-varying verifier) and baselines, from a 6-rollout pilot. (b)~Extrapolation error by
  threshold. (c)~Choosing the threshold for each number of rollouts on unseen issues. Argmax of the
  verifier score (pink) matches the best threshold at $N=1$ and exceeds every threshold for $N\ge2$.}
  \label{fig:swe}
\end{figure}

\Cref{tab:regret-sampling,tab:regret-verify} summarise the main result (see also
\cref{fig:alloc-real}, and \cref{fig:alloc-sim,fig:phase} in the appendix). In every setting, a
variant of SPLIT is within 0.8 points of the hindsight-optimal configuration on
tasks it never saw, and the best variant is within 0.02--0.65 points. Every fixed rule fails
somewhere, for an identifiable reason:
\begin{itemize}[leftmargin=*,itemsep=1pt,topsep=1pt]
  \item \textbf{``Strongest model'' and the Singhi et al.\ rule} are near-optimal at large budgets
  but lose about 12 points on MATH. At small budgets a single Llama-3-70B sample (cost 7.8--8.75)
  is unaffordable, so they submit nothing. The planner instead draws three Llama-3-8B samples and
  reaches 36\% at a budget of 4. The Singhi rule is slightly better than the planner on
  CodeContests with parameter-count costs (0.78 vs 0.82) and on MATH-500 (0.05 vs 0.06), but loses
  under compute-matched costs (0.73 vs 0.65) and on GPQA (1.20 vs 0.02).
  \item \textbf{``Generation only''} is right for MATH-500, where extra judge calls are useless, and
  loses 2.6 points on GPQA, where they help. \textbf{``Maximum verification''} is never best.
  \item \textbf{A fixed verifier threshold} is wrong at both ends of the budget range on SWE-bench
  (\cref{fig:swe}c). With one rollout, filtering through the verifier at threshold 0.5
  \emph{lowers} the resolve rate from 32.5\% to 25.1\%, because it rejects correct patches with no
  second attempt to fall back on. The planner uses no verifier at $N=1$, a lenient threshold at
  $N=2$ and a stricter one from $N=4$.
\end{itemize}

Modelling correlation matters for decisions. Assuming independent errors raises regret 1.2--4.8
times (e.g.\ 0.54 to 1.69 on MATH, 0.11 to 0.53 on GPQA; paired $p \le 0.003$ on MATH,
CodeContests and real tests). Modelling task-dependent verifier accuracy matters on agents: the
pooled-verifier variant mis-picks the strictest threshold at $N=16$ and has regret 2.36, against 0.33
with the task-varying model ($p=0.005$). The automatic choice between the two verifier models
gives 0.56, a safe default when the right model is unknown (\cref{tab:auto}). Results are unchanged in kind under
compute-matched costs, where a Llama-3-70B sample costs 7.8 (MATH) or 10.7 (CodeContests)
Llama-3-8B samples. Two comparisons are not significant: the planner against ``strongest model''
with real CodeContests tests ($-0.32$ points, $p=0.43$), and against the Singhi rule averaged over
the two simulated datasets ($p=0.13$ and $0.09$ under compute-matched costs), where a large win on
MATH is paired with a near tie on CodeContests.

\begin{figure}[t]
  \centering
  \includegraphics[width=\linewidth]{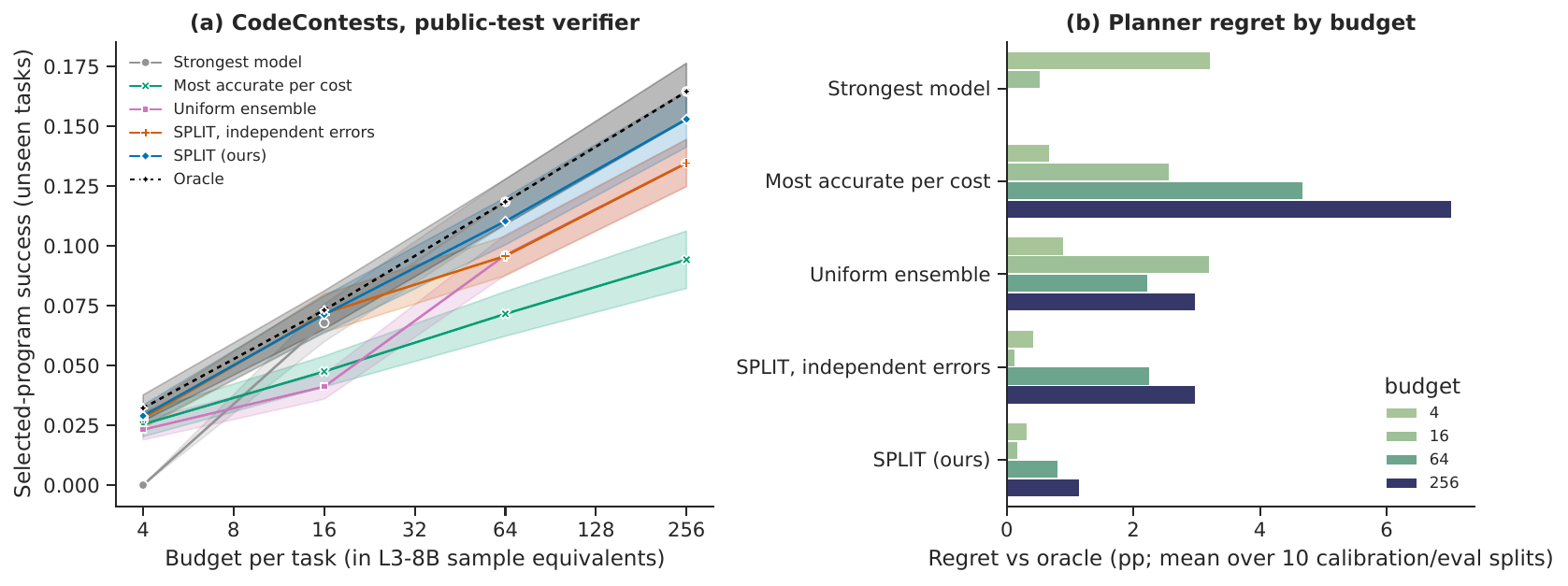}
  \caption{\textbf{Allocation with a real verifier} (CodeContests public tests, planner calibrated on
  50\% of tasks with 32-sample pilots, 10 splits). (a)~Success on held-out tasks of the
  configuration chosen by each method against the per-task budget. (b)~Regret relative to the
  oracle at each budget. SPLIT (ours) stays within about 1.2 points of the oracle at every
  budget; each fixed rule is good only in part of the range.}
  \label{fig:alloc-real}
\end{figure}

\subsection{Why regret is small, other costs, and online re-planning}
\label{sec:res-beyond}

\paragraph{Regret tracks differential, not absolute, error.} \Cref{prop:regret}(a) bounds regret by
the difference between the prediction errors of the chosen and the optimal configuration. On
CodeContests with real tests, the largest absolute prediction error over the 13 configurations
averages 3.2--8.5 points across budgets, but this pairwise error averages only 0.6--2.8 points and
the regret 0.2--1.2 (\cref{tab:regret-diag}). On MATH with a simulated verifier the gap is wider
still: the largest absolute error averages 11.1 points, the regret 0.54. Most of the absolute error is shared by all
configurations and comes from the task sample: the calibration and evaluation halves differ by up
to 3--8 points in true success, while SPLIT's own error on the calibration tasks varies by only
0.6--4.9 points across configurations. The finite-sample term of \cref{prop:regret}(c) is
vacuous here (60 points at $T_{\mathrm{cal}}=T_{\mathrm{ev}}=70$, $|\mathcal C|=13$, $\delta=0.05$), so
the guarantee explains consistency and the $\log|\mathcal C|$ dependence, not the size of the
observed regret.

\begin{figure}[t]
  \centering
  \includegraphics[width=\linewidth]{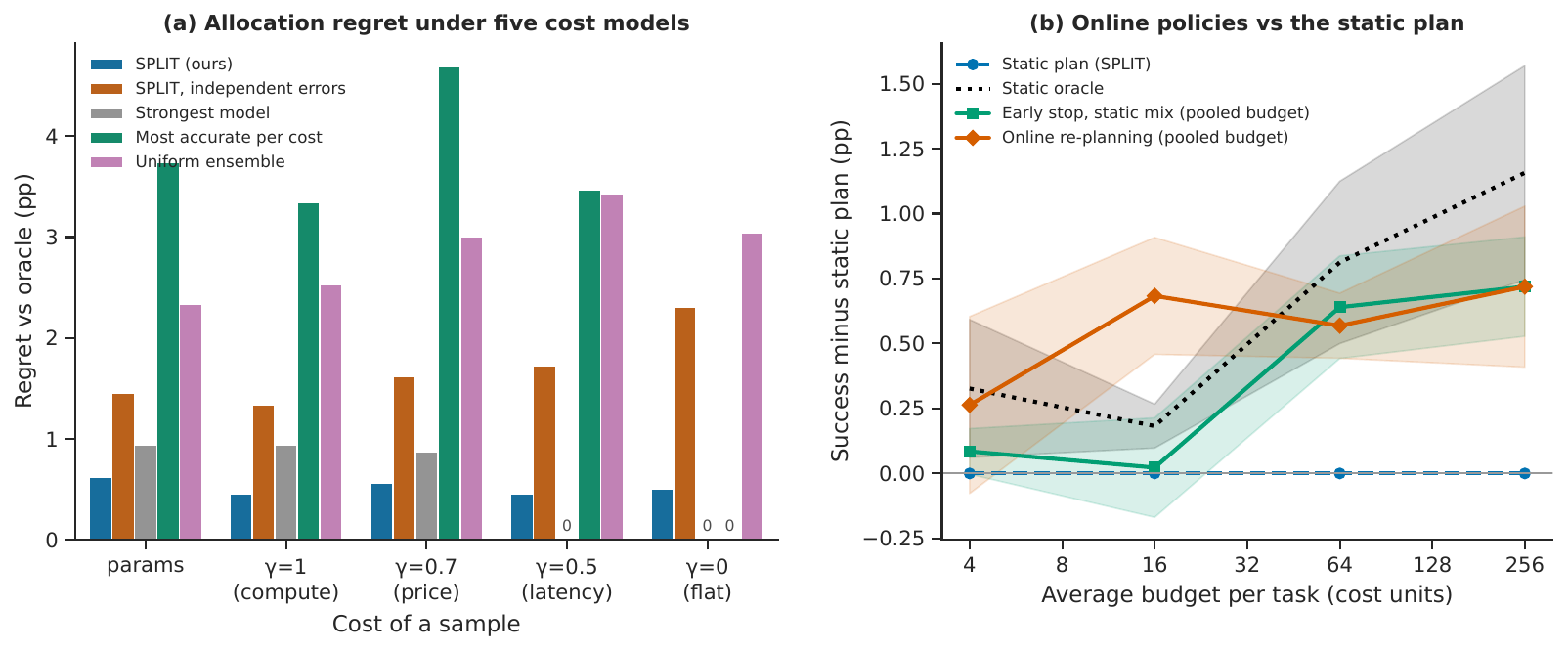}
  \caption{\textbf{Cost models and online re-planning} (CodeContests, public tests as the
  verifier, 10 splits). (a)~Regret of each allocation rule when a sample costs
  $(\text{parameters})^\gamma\times$ output tokens; ``params'' prices by parameters only. Zero bars
  are labelled. (b)~Success of online policies minus the static plan (mean $\pm$ s.e.\ over
  splits). Budgets are pooled across the evaluation tasks.}
  \label{fig:prices-online}
\end{figure}

\paragraph{Other prices.} Which fixed rule is right depends on how a sample is priced
(\cref{fig:prices-online}a). On CodeContests with real tests, ``strongest model'' is optimal when
per-token prices are flat or grow with the square root of model size ($\gamma\le0.5$), and loses
0.9 points otherwise. ``Most accurate per cost'' is optimal under flat prices and loses 3.3--4.7
points under every other price. The planner's regret stays between 0.45 and 0.62 points under all
five cost models. It is never worse than 0.62, but under flat or latency-like prices, where one
model dominates, a fixed rule that happens to pick that model beats it by about half a point
(0.45--0.50, $p\le0.001$). With
the simulated verifier under $\gamma=0.7$ prices, the planner's regret is 0.16 points on MATH and
0.79 on CodeContests, against 10.44 and 1.06 for ``strongest model'' and 9.69 and 0.68 for the
Singhi rule. This repeats the pattern of \cref{tab:regret-sampling}: a large win on MATH and a near
tie on CodeContests.

\begin{table}[t]
\centering
\caption{\textbf{Online re-planning on unseen CodeContests tasks} (success in \%, mean over 10 splits;
16 Monte Carlo replays per split). ``Spent'' is the fraction of the budget used. Policies with a
per-task budget stop a task at its first accepted candidate. Policies with a pooled budget give
the savings to unresolved tasks.}
\label{tab:online}
\small
\setlength{\tabcolsep}{5pt}
\begin{tabular}{@{}lcccccccc@{}}
\toprule
& \multicolumn{2}{c}{$B=4$} & \multicolumn{2}{c}{$B=16$} & \multicolumn{2}{c}{$B=64$} & \multicolumn{2}{c}{$B=256$} \\
\cmidrule(lr){2-3}\cmidrule(lr){4-5}\cmidrule(lr){6-7}\cmidrule(lr){8-9}
Policy & success & spent & success & spent & success & spent & success & spent \\
\midrule
Static plan (SPLIT) & 2.90 & 1.00 & 7.14 & 1.00 & 11.04 & 1.00 & 15.31 & 1.00 \\
Static oracle (hindsight) & 3.22 & 1.00 & 7.32 & 1.00 & \textbf{11.85} & 1.00 & \textbf{16.47} & 1.00 \\
\midrule
Early stop, static mix, per task & 2.82 & 0.95 & 6.90 & 0.93 & 10.91 & 0.85 & 15.21 & 0.75 \\
Re-planning, per task & 2.79 & 0.95 & 7.42 & 0.93 & 11.09 & 0.86 & 15.13 & 0.75 \\
Early stop, static mix, pooled & 2.98 & 1.00 & 7.16 & 1.00 & 11.68 & 1.00 & 16.03 & 1.00 \\
\textbf{Re-planning, pooled} & \textbf{3.16} & 1.00 & \textbf{7.82} & 1.00 & 11.61 & 1.00 & 16.03 & 1.00 \\
\bottomrule
\end{tabular}
\end{table}

\paragraph{Online re-planning.} The planner fixes a configuration before seeing any task. Under
uniform-among-accepted selection, once one candidate of a task is accepted, its success equals the
precision and further samples cannot raise it. So a policy can stop a task at its first acceptance
and spend the remainder elsewhere. We evaluate two online policies on the E7 splits, simulating
samples from each evaluation task's measured correctness and verifier rates, the same rates that
score the static plan (\cref{tab:online,fig:prices-online}b). \emph{Early stopping} keeps the static
plan's generator mix. \emph{Re-planning} keeps a per-task posterior over the generators' rates
(importance weights on the copula prior draws, updated by each rejection) and draws the generator,
and with a pooled budget the task, with the highest posterior probability of a correct acceptance
per unit cost. With a per-task budget, stopping at the first acceptance reaches the static plan's
success while spending 5--25\% less. Pooling the savings raises success above the static plan by
0.56 points on average (95\% bootstrap interval 0.30--0.80; Wilcoxon $p<10^{-5}$; better in 33 of
40 split--budget pairs). This is statistically indistinguishable from the best \emph{static}
configuration in hindsight ($-0.06$ points, $p=0.49$), which re-planning beats at $B=16$. Most of
the gain comes from early stopping with a pooled budget. Per-task generator switching adds to it
only at small budgets (0.66 points at $B=16$); averaged over budgets, re-planning and early
stopping with the static mix differ by 0.19 points, which is not significant ($p=0.28$). The gains
are modest in absolute terms because most CodeContests
tasks are never solved by any generator and the public tests rarely accept wrong programs, which
leaves little to re-allocate.

\subsection{Mechanisms}
\label{sec:res-mech}

\begin{figure}[t]
  \centering
  \includegraphics[width=\linewidth]{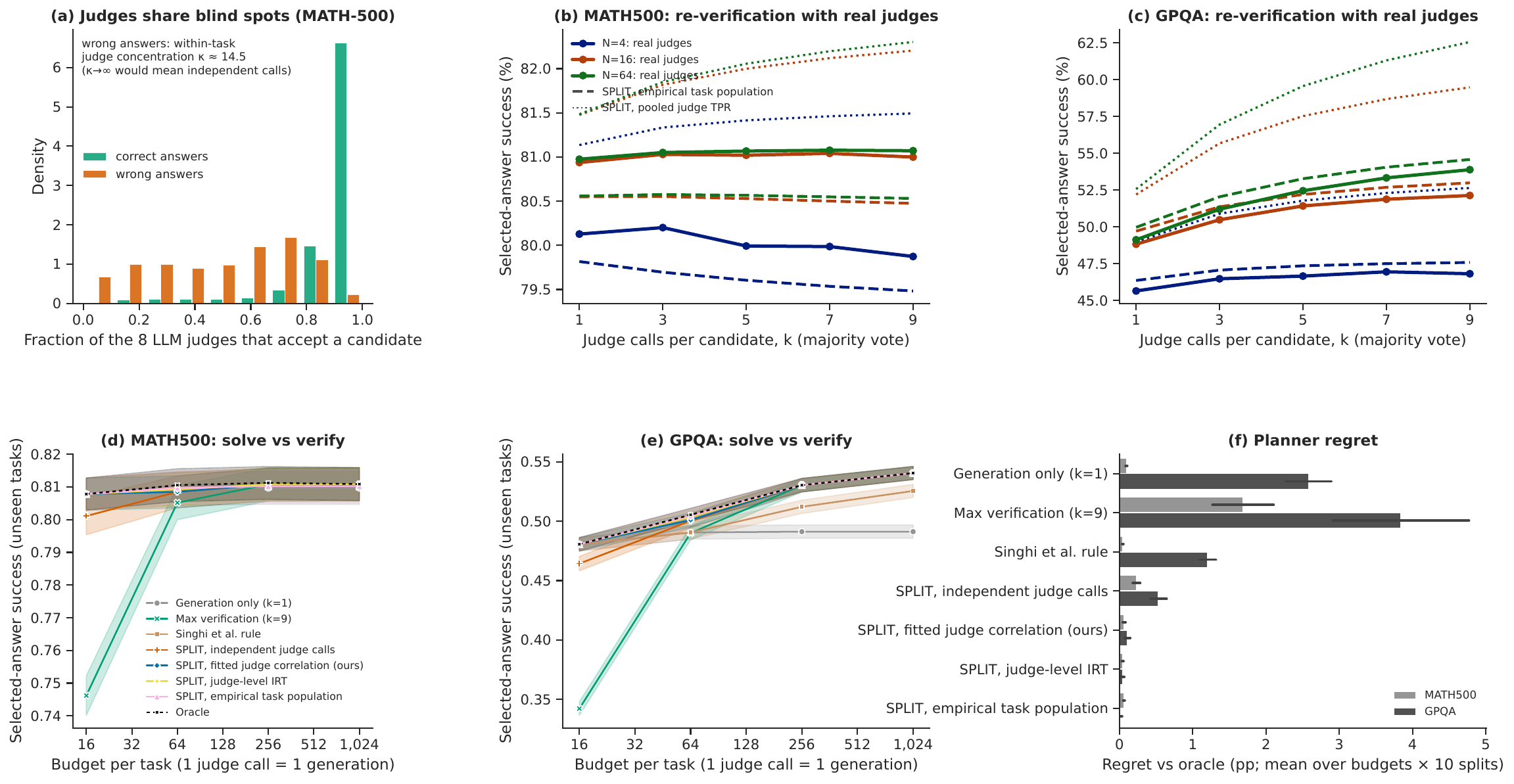}
  \caption{\textbf{Stochastic LLM-judge verification} (Weaver release, Llama-3.1-70B-Instruct
  candidates). (a)~Many wrong MATH-500 answers are accepted by most of the eight judges. (b,~c)~Verified
  success against the number of judge calls $k$ for $N=4,16,64$: real (solid), SPLIT with the
  empirical task population (dashed), and SPLIT with pooled judge rates (dotted). Extra calls are
  useless on MATH-500 and help on GPQA. (d--f)~Solve-vs-verify allocation on unseen tasks and
  planner regret.}
  \label{fig:judges}
\end{figure}

\paragraph{Re-verification lowers the precision ceiling exactly when \cref{eq:hurt} holds.} We treat a
call to a random judge among the eight LLM judges in the Weaver release as a stochastic verifier and
take the majority of $k$ calls. Pooled over MATH-500 candidates, going from $k=1$ to $k=9$ multiplies
the acceptance of wrong answers by $1.052$ ($0.507\to0.534$) but the acceptance of correct answers by
only $1.026$ ($0.931\to0.955$). Because pooled ratios mix heterogeneous tasks, we also evaluate the
condition task by task, using each task's own candidates and judge propensities. Nine calls lower the
precision ceiling on 61\% of MATH-500 tasks, and the mean per-task ceiling is flat (0.717 vs 0.718).
Measured verified success changes by only $-0.3$ to $+0.1$ points across $N$ (\cref{fig:judges}b).
Many wrong MATH answers fool most judges (\cref{fig:judges}a): a third of the 11{,}015 wrong
answers are accepted by at least six of the eight judges. Majority voting therefore amplifies the
error. On GPQA, pooled wrong-answer acceptance \emph{falls} ($\times0.75$) while correct-answer
acceptance rises ($\times1.07$). Nine calls raise the ceiling on 86\% of tasks (mean 0.569 to 0.638)
and add $+1.2$, $+3.3$ and $+4.8$ points of verified success at $N=4$, 16 and 64
(\cref{fig:judges}c).

\paragraph{Pooling verifier accuracy across tasks biases forecasts.} SPLIT with pooled judge rates
gets the direction of these effects right but overstates the GPQA gain about twofold
(\cref{tab:reverify}). The cause is not the difficulty prior: a non-parametric prior does no
better, and with each task's true rates SPLIT matches simulation to 0.2 points. The cause is that
judge accuracy is coupled to task difficulty. On GPQA, the per-task TPR and FPR of the judge pool
have Spearman correlations of 0.54 and 0.65 with task difficulty, and a pooled rate erases this.
Averaging the formula over the calibration tasks' own candidates and judge propensities (the
\emph{empirical task population}) removes the bias. It predicts the gain at every $N$ to within
0.2 points (\cref{tab:reverify}) and gives the lowest allocation regret on GPQA (0.02 points). A
judge-level IRT model improves decisions but not the size of the effect. The same coupling
appears on agents. An LLM patch verifier is more lenient on easy issues, and modelling per-issue
verifier rates halves the forecast error on SWE-bench (5.35 to 2.74 points; \cref{tab:forecast}).

\begin{table}[t]
\centering
\caption{\textbf{Gain from 9 judge calls instead of 1} on GPQA (percentage points of verified
success), real versus predicted from a 32-sample pilot on calibration tasks.}
\label{tab:reverify}
\small
\begin{tabular}{@{}lccc@{}}
\toprule
& $N=4$ & $N=16$ & $N=64$ \\
\midrule
Real & $+1.2$ & $+3.3$ & $+4.8$ \\
\textbf{SPLIT, empirical task population} & $\mathbf{+1.2}$ & $\mathbf{+3.3}$ & $\mathbf{+4.6}$ \\
SPLIT, pooled judge rates & $+3.7$ & $+7.3$ & $+10.0$ \\
SPLIT, judge-level IRT & $+3.4$ & $+6.6$ & $+9.6$ \\
\bottomrule
\end{tabular}
\end{table}

\paragraph{A strong scorer is not a universal selector.} Submitting the candidate with the highest
Weaver aggregated score \citep{saadfalcon2025weaver} reaches 86--93\% on MATH-500 ($N=4$ to 64), well
above judge-based selection ($\approx$81\%). On GPQA it reaches only $\approx$43\%, below a single
judge call (46--49\%). Which selector to use is itself a decision that a pilot can make.

\begin{figure}[t]
  \centering
  \includegraphics[width=\linewidth]{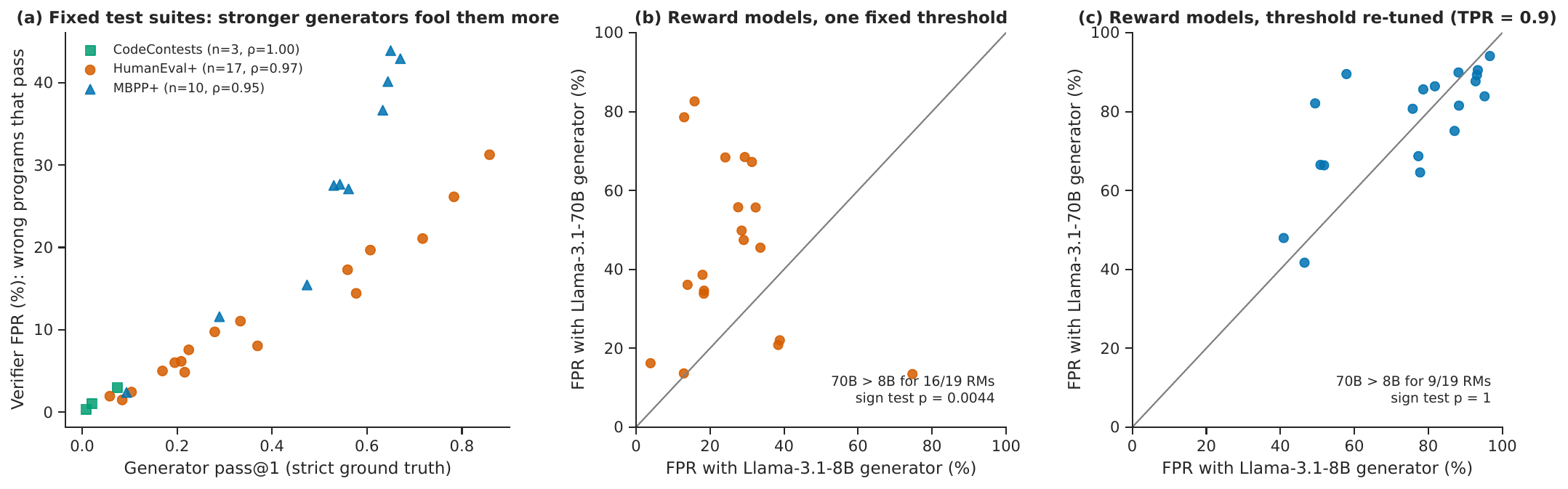}
  \caption{\textbf{Capability and verifier false positives.} (a)~For fixed test suites, the FPR rises
  with generator pass@1 on HumanEval+, MBPP+ and CodeContests (replicating
  \citealp{stroebl2024limits}). (b,~c)~For 19 reward models, the stronger generator's wrong answers
  pass a fixed threshold more often (b), but not once the threshold is re-tuned to equal TPR (c).}
  \label{fig:capfpr}
\end{figure}

\paragraph{Stronger generators fool fixed tests more; for scored verifiers this is calibration.}
\citet{stroebl2024limits} reported that a model's single-sample accuracy is strongly correlated
with its false-positive rate on HumanEval and MBPP. We replicate this with per-task FPRs on the
HumanEval+ ground truth: Spearman $\rho=0.97$ across 17 models ($p=6\times10^{-11}$), from 1.5\% for
Vicuna-7B to 31\% for GPT-4o. We also replicate it on MBPP+ ($\rho=0.95$ across 10 models,
$p=2\times10^{-5}$), and find the same ordering for CodeContests public tests (\cref{fig:capfpr}a).
For \emph{scored} verifiers the picture differs. Across 19 reward models on MATH-500, the wrong
answers of Llama-3.1-70B-Instruct pass a fixed threshold more often than those of
Llama-3.1-8B-Instruct for 16 of 19 reward models (sign test $p=0.004$). When the threshold is re-tuned
so that each generator's correct answers pass at the same 90\% rate, the effect disappears (9 of
19, $p=1$). The capability effect for reward models is a shift of the whole score distribution,
which per-generator calibration removes; pass/fail tests offer no such knob. For allocation, this
means the verifier must be re-measured whenever the generator changes, which the pilot does.

\paragraph{Shared blind spots set the value of mixing models.} Generators mostly fail on the same
problems. On MATH, the per-problem solve rates of Llama-3-8B-Instruct and Gemma-7B correlate at
0.80, and 38 of the 128 problems are solved less than 5\% of the time by both. Pilot-estimated copula correlations between model pairs are 0.67--0.96 on MATH and
CodeContests (\cref{fig:diversity}, \cref{app:figures}). Assuming independent failures overstates
the gain from mixing two models by up to about 7 points, with a mean error of 2.81 points, whereas
SPLIT (with the copula) predicts the gain with a mean error of 0.43 points. For Llama-3-8B-Instruct and
Gemma-7B on MATH ($\hat\rho=0.83$), independence promises $+5$ points; the real gain is $+1.6$ and the
SPLIT predicts $+2.4$. The same structure appears among agents: of 500 SWE-Bench Verified issues, 26
are resolved by none of 175 submitted systems and 67 by fewer than 5\% of them (\cref{fig:pool}d).

\subsection{Forecasting success far beyond the pilot}
\label{sec:res-forecast}

\begin{table}[t]
\centering
\caption{\textbf{Extrapolation error} (MAE in percentage points over all $N$ beyond the pilot;
mean over pilot sizes and seeds unless stated). ``Ours'' is the best-justified variant per row:
pooled verifier for a simulated verifier and CodeContests; task-varying verifier for HumanEval
and SWE-bench thresholds; issue-level score shift for argmax; NPMLE prior for the agent pool.
Paired $p$-values compare ours with the power law.}
\label{tab:forecast}
\small
\setlength{\tabcolsep}{4pt}
\begin{tabular}{@{}llrcccc@{}}
\toprule
Setting & Target & Horizon & Ours & Power law & Plug-in & $p$ \\
\midrule
MATH + CodeContests, sim.\ verifier (pilot 32) & selection & 8{,}192 & \textbf{1.9} & 7.6 & --- & $<10^{-3}$ \\
MATH + CodeContests (pilot 32) & coverage & 8{,}192 & 3.3 & \textbf{2.6} & --- & --- \\
CodeContests, public tests & selection & 8{,}192 & \textbf{2.13} & 2.77 & 5.54 & 0.002 \\
HumanEval, base tests (17 models) & selection & $\approx$128 & \textbf{1.69} & 2.13 & 4.55 & $<10^{-3}$ \\
SWE-bench, LLM verifier thresholds & selection & 16 & \textbf{2.74} & 6.70 & 4.07 & $<10^{-3}$ \\
SWE-bench, argmax of verifier score & selection & 16 & 3.28 & \textbf{1.58} & 2.11 & 0.006 \\
SWE-bench agent pool (175 systems) & coverage & 25--167 & 2.23 & \textbf{1.94} & 4.75 & 0.57 \\
\bottomrule
\end{tabular}
\end{table}

The allocation results need SPLIT to rank configurations correctly, not to forecast exact
success rates. \Cref{tab:forecast} reports how well it does as a forecaster. The picture is mixed,
and we report it in full.

\begin{figure}[t]
  \centering
  \includegraphics[width=\linewidth]{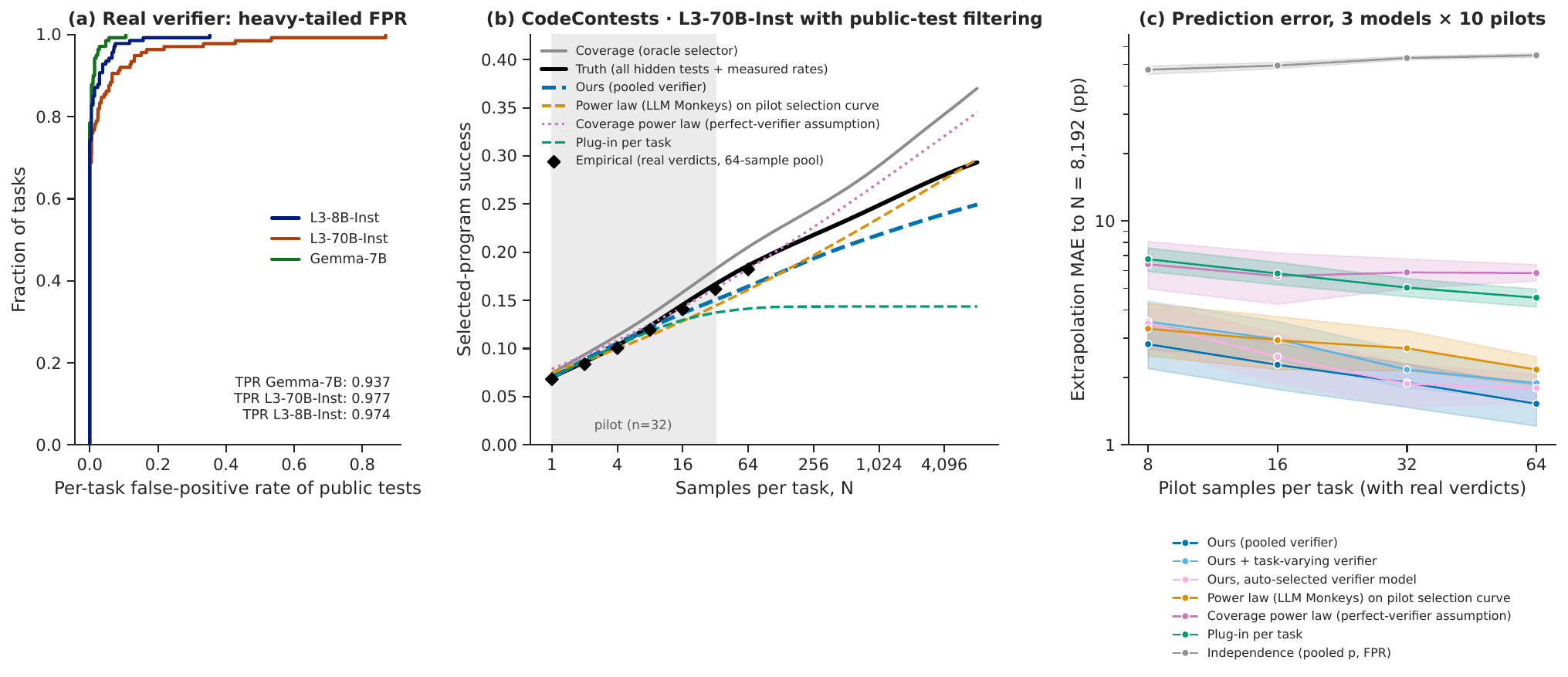}
  \caption{\textbf{CodeContests public tests as a real verifier.} (a)~Per-task FPR is zero for most
  tasks but heavy-tailed. (b)~Truth (all tests) versus predictions from a 32-sample pilot.
  (c)~Extrapolation error to $N=8{,}192$ against pilot size for three generators and 10 pilots.}
  \label{fig:cc-real}
\end{figure}

\paragraph{Where SPLIT helps.} With a verifier that has a non-zero FPR, success plateaus. A power
law fitted to the pilot keeps predicting growth and misses by 7.6 points at 8{,}192 samples; SPLIT
misses by 1.9 (simulated verifier; \cref{fig:pilot-sim}). The power law fitted to coverage, which implicitly assumes a
perfect verifier, is off by 6--24 points with real verifiers. With real tests the margin over the
power law is small but significant: 0.6 points on CodeContests (80{,}403 executed programs,
\cref{fig:cc-real}) and 0.4 points on HumanEval across 17 generators (\cref{fig:humaneval}). On agentic SWE-bench the
task-varying verifier model is 4 points better than the power law.

\paragraph{Where it does not.} For pure coverage, the power law is slightly better than the
beta-binomial prior in SPLIT (2.6 vs 3.3 points), consistent with the heavy-tailed explanation of
\citet{schaeffer2025monkeys}. For argmax selection on SWE-bench, our estimator improves from 4.5
to 3.3 points with the per-issue score shift, and its error at $N=16$ ($+2.4$) matches the power
law's ($+2.3$). Averaged over $N$, the power law is still better (1.6 points, $p=0.006$;
\cref{tab:argmax}, \cref{fig:argmax}). The
estimator is sensitive to the upper tail of the score distribution, as \cref{prop:argmax}
predicts: replacing the empirical residual tails with Gaussian tails turns a 5-point
over-prediction into a 7-point under-prediction.

\begin{figure}[t]
  \centering
  \includegraphics[width=\linewidth]{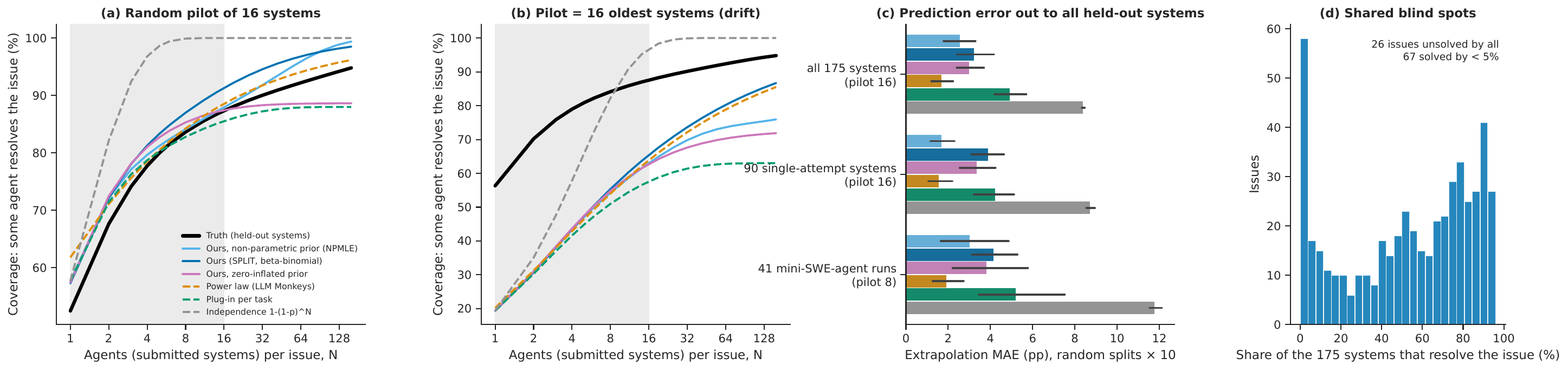}
  \caption{\textbf{A pool of 175 SWE-Bench Verified submissions as attempts} (coverage only; no
  shared verifier). (a)~Random pilot of 16 systems; (b)~pilot of the 16 oldest systems (capability
  drift); (c)~extrapolation error over 10 random splits for three pools; (d)~share of systems that
  resolve each issue. Issues no system resolves cap coverage.}
  \label{fig:pool}
\end{figure}

\paragraph{A long horizon with real agents.} No public release we found has more than about 25
labelled rollouts of one agent per SWE-bench issue. As a long-horizon test of the
\emph{coverage} component (the difficulty prior), we treat an ``attempt'' as a system drawn at random
from the 175 SWE-bench Verified submissions. Random splitting makes systems exchangeable, so for each
issue the held-out systems behave as draws with a common success rate $p_t$ and the coverage term of the formula
applies. There is no shared verifier, so this does not test selection, and the attempts come from
different policies rather than repeated runs of one agent. We calibrate on a random pilot of 4--32
systems and predict coverage out to every held-out system, up to about 20 times the pilot for the
smallest pilots (\cref{fig:pool}, \cref{tab:pool}). The
power law and SPLIT with an NPMLE prior are statistically tied (1.9 vs 2.2 points, $p=0.57$). The
beta prior is worse (3.8 points) because it puts too little mass on unsolvable issues. With the NPMLE prior,
SPLIT is nearly unbiased at the largest $N$: its mean signed error is $-2.3$ to $+0.1$ points, while
the power law over-predicts in every configuration ($+0.5$ to $+3.6$). It is noisier from pilot to
pilot, however, so its mean absolute end-point error is larger (2.9 vs 2.0, $p=0.004$). When the
pilot consists of the \emph{oldest} systems, every method under-predicts badly for small pilots,
because later systems are much stronger. A pilot must represent the generators that will actually
be deployed.

\section{Limitations and conclusion}
\label{sec:limitations}

\paragraph{Limitations.}
\begin{itemize}[leftmargin=*,itemsep=1pt,topsep=1pt]
  \item \textbf{Point forecasts.} SPLIT's advantage over power-law extrapolation is large only
  under a simulated verifier and on SWE-bench thresholds. It is small (0.4--0.6 points) with real
  tests, absent for pure coverage, and negative for argmax selection (\cref{tab:forecast}). Our
  claims rest on decisions and mechanisms, not on forecasting accuracy.
  \item \textbf{Same-policy agent horizons.} The longest same-agent horizon we can test is 16
  held-out rollouts per issue, because no public release we found has more. The 175-system pool
  tests long horizons for coverage only, with no shared verifier. Testing selection at long
  horizons with one agent needs new rollouts. We release a pipeline that collects, scores and
  analyses them (\cref{app:repro}), but have not run it: at 64 rollouts per issue it needs about
  32{,}000 agent runs.
  \item \textbf{Representative pilots.} All estimators assume the pilot is exchangeable with
  deployment. Under capability drift (pilot of older systems), every method fails for small
  pilots (\cref{sec:res-forecast}). Some refinements also need labelled calibration tasks with
  repeated judge verdicts.
  \item \textbf{Costs.} Prices and latency enter through a one-parameter family,
  $(\text{parameters})^\gamma\times$ measured tokens, not through quoted provider prices. Latency is
  modelled as additive cost, which ignores parallel sampling. Judge calls are priced as one
  generation.
  \item \textbf{Model choice.} Neither the pooled nor the task-varying verifier model is best on every
  dataset. Our out-of-sample selection rule is never the worse of the two but is significantly
  behind the better one on two of three datasets (\cref{app:extra}).
  \item \textbf{Theory.} The propositions are exact under conditional independence of candidates
  given the task (\cref{ass:ci}). Dependence among candidates \emph{within} a task, such as
  duplicated or near-duplicated answers, is not modelled and may bias SPLIT, especially at large
  $N$. The verifier propensity models repeated calls on one candidate, not correlated candidates. The
  planner's finite-sample regret guarantee (\cref{prop:regret}) is vacuous at our task counts; it
  explains consistency, not the size of the observed regret.
  \item \textbf{Online policy.} The re-planning policy is a greedy index, not an optimal sequential
  policy. It ignores the risk that a false acceptance ends a task, and is evaluated only on
  CodeContests, by simulation from measured per-task rates.
  \item \textbf{Agent pool.} The 175-system experiment treats different systems as exchangeable
  attempts. This tests the difficulty prior at long horizons but is not a test of one agent's
  repeated sampling or of selection.
\end{itemize}

\paragraph{Conclusion.} Test-time compute is limited by selection as much as by generation. A
small labelled pilot, interpreted through an exact success formula that accounts for difficulty,
shared blind spots and correlated verifier errors, is enough to decide how many samples to draw,
which generators to mix, and how much to verify. Across math, science QA, code and agents this
decision is within a fraction of a point of the best possible, where fixed rules can lose 2.5 to 20
points. The planner needs only to rank configurations, so its regret is bounded by differential,
not absolute, forecast error, and it stays near-optimal when samples are priced by compute, by
sublinear per-token prices or by latency. The same model explains when re-verification backfires and
why pooled verifier statistics mislead. Re-planning online, by stopping a task at its first
accepted candidate and moving the savings to unresolved tasks, recovers the gap to the best static
configuration in hindsight. The most useful next step is to test selection on long same-agent
rollouts, for which we release a ready-to-run pipeline.

\subsubsection*{Broader Impact Statement}
This work helps practitioners spend inference compute more efficiently, which can reduce the
cost and energy use of deployed language-model systems. It also documents failure modes that
matter for oversight. Stronger generators produce wrong outputs that pass fixed checks more often,
and asking more LLM judges can make acceptance of wrong answers \emph{more} likely when judges share
blind spots. Systems that rely on automated verification for safety-relevant decisions should
re-measure verifier error rates whenever the generator changes, and should not assume that
majority votes of judges converge to the truth. Our planner optimises measured success on a
benchmark-defined notion of correctness; when that notion is itself flawed (e.g.\ weak tests), the
planner will optimise the flawed target. All data are public releases; we collected no human data.

\bibliography{references}
\bibliographystyle{tmlr}

\clearpage
\appendix
\section{Proofs}
\label{app:proofs}

Throughout, \cref{ass:ci} holds, abstention counts as failure, $a/s := 0$ when $s=0$, and
$0^0 = 1$. Numerical checks of every statement below, of the corrected conditions and of the edge
cases (26 checks) are included in the code release.

\subsection{Proposition~\ref{prop:law} (exact success formula)}

\begin{proof}
Give each candidate an independent arrival time $U_i \sim \mathrm{Unif}(0,1)$, independent of all
$(C_j, A_j)$. Choosing uniformly among accepted candidates has the same distribution as choosing the
accepted candidate with the smallest $U_i$, because the arrival order is a uniformly random
permutation independent of acceptance. Candidate $i$ is selected and correct iff it is accepted
and correct and every $j$ with $U_j < U_i$ is rejected. Conditioning on $U_i=u$ and using
independence across candidates,
\[
  \Prob(i \text{ selected and correct} \mid U_i = u) = a_i \prod_{j\ne i} (1 - s_j u),
\]
since $\Prob(A_j = 1, U_j < u) = s_j u$. Integrating over $u$ and summing over $i$, grouping the
$n_m$ exchangeable candidates of generator $m$, gives \cref{eq:law}. For one generator,
$\int_0^1 N a (1-su)^{N-1} du = \frac{a}{s}(1-(1-s)^N)$ for $s>0$, which is non-decreasing in $N$
with limit $a/s$. If $s=0$ then $a=0$ and $S_N=0$.
\end{proof}

The implementation uses the closed form for one generator and evaluates the mixed-generator
integral by 480-node log-spaced trapezoidal quadrature; it agrees with Monte Carlo to within
$3\times10^{-3}$.

\subsection{Convergence regimes}

\begin{proposition}[Regimes]\label{prop:regimes}
Let $p \sim \pi$ on $[0,1]$, $\tau \in (0,1]$, one generator, and $S_\infty = \E[a/s]$.
\begin{enumerate}[label=(\alph*),leftmargin=*,itemsep=1pt]
  \item If $f_t \ge f_0 > 0$ for all tasks, then $0 \le S_\infty - S_N \le S_\infty (1 - \min(\tau, f_0))^N
  \le S_\infty e^{-N\min(\tau,f_0)}$.
  \item If $\pi$ has a density with $\pi(p)/(c\,p^{\alpha-1}) \to 1$ as $p\to 0$ for some $c,\alpha>0$, then
  $\E[(1-p)^N] \sim c\,\Gamma(\alpha) N^{-\alpha}$.
  \item If $\Prob(f_t \le x) \le K x^{a_f}$ for $0 \le x \le x_0$, with $a_f, x_0 > 0$, then
  $S_\infty - S_N \le (1-\tau)^N + K\Gamma(a_f+1) N^{-a_f} + e^{-N x_0}$. This is an upper bound only.
  \item For any FPRs $f_t \ge 0$ and any $N$, $0 \le \E[1-(1-\tau p)^N] - S_N \le N\,\E[f_t(1-p)] \le N\,\E[f_t]$.
\end{enumerate}
\end{proposition}

\begin{proof}
(a)~$S_\infty - S_N = \E[(a/s)(1-s)^N]$, and $s = \tau p + f(1-p)$ is a convex combination of $\tau$
and $f$, so $s \ge \min(\tau, f_0)$; then use $1-x\le e^{-x}$.
(b)~Fix $\delta$ with $\pi(p) \le 2cp^{\alpha-1}$ on $(0,\delta)$. The integral over $[\delta,1]$ is at
most $(1-\delta)^N$. On $(0,\delta)$, substitute $p=x/N$:
$N^\alpha \int_0^\delta (1-p)^N \pi(p)\,dp = \int_0^{N\delta} (1-x/N)^N N^{\alpha-1}\pi(x/N)\,dx$. The integrand converges
to $c\,x^{\alpha-1}e^{-x}$ and is dominated by $2c\,x^{\alpha-1}e^{-x}$, so dominated convergence gives
$c\,\Gamma(\alpha)$ \citep[cf.][]{schaeffer2025monkeys}.
(c)~As in (a), $a/s \le 1$ and $s_t \ge \min(\tau, f_t)$, so $S_\infty - S_N \le (1-\tau)^N + \E[e^{-Nf_t}]$.
Writing the expectation through the CDF,
$\E[e^{-Nf_t}] = \int_0^\infty N e^{-Nx}\Prob(f_t \le x)\,dx \le \int_0^{x_0} N e^{-Nx} K x^{a_f} dx +
\int_{x_0}^\infty N e^{-Nx} dx \le K\Gamma(a_f+1)N^{-a_f} + e^{-Nx_0}$. The true gap can decay faster
(for instance if tasks with small $f_t$ also have small $p$).
(d)~Couple the verifier with FPR $f_t$ to one with FPR $0$: same labels, same acceptance of correct
candidates, same arrival times, and wrong candidates accepted only by the first verifier. If no
wrong candidate is accepted, both selectors make the same choice. The FPR-0 selector succeeds iff
some correct candidate is accepted, with probability $1-(1-\tau p)^N$. The other can only do worse,
and only on the event that a wrong candidate is accepted, which has probability at most
$N f_t (1-p)$ by a union bound.
\end{proof}

\subsection{Shared blind spots}

\begin{proposition}[Diversity]\label{prop:diversity}
Let two generators have per-task rates $(p_A,p_B)$ with $p_A,p_B<1$ a.s., conditionally independent
labels given $(p_A,p_B)$, a perfect verifier, a budget $N>0$ split as $(1-\lambda)N$ and $\lambda N$
($\lambda\in[0,1]$ continuous), and $F(\lambda) = 1-\E[(1-p_A)^{(1-\lambda)N}(1-p_B)^{\lambda N}]$.
Assume the weighted log-moments below are finite.
\begin{enumerate}[label=(\alph*),leftmargin=*,itemsep=1pt]
  \item $F$ is concave.
  \item $F(\lambda) > F(0)$ for some $\lambda>0$ iff $\E[w_A\log(1-p_B)] < \E[w_A \log(1-p_A)]$ with
  $w_A = (1-p_A)^N$. An interior $\lambda$ strictly better than both endpoints exists iff
  $F'_+(0) > 0$ and $F'_-(1) < 0$.
  \item If $p_A$ and $p_B$ have the same marginal law, $F(\tfrac12) \ge F(0)$, with equality iff
  $p_A = p_B$ a.s.
  \item If $p_A = F_A^{-1}(\Phi(Z_A))$, $p_B = F_B^{-1}(\Phi(Z_B))$ with $(Z_A,Z_B)$ standard bivariate
  normal with correlation $\rho$, then $F(\tfrac12) - F(0)$ is non-increasing in $\rho\in[-1,1]$.
\end{enumerate}
\end{proposition}

\begin{proof}
(a)~With $\ell_A = \log(1-p_A)$ and $\ell_B = \log(1-p_B)$, $1-F(\lambda) = \E[\exp(N((1-\lambda)\ell_A
+ \lambda\ell_B))]$ is an expectation of the exponential of an affine function of $\lambda$, hence
convex.
(b)~$F'_+(0) = -N\,\E[(1-p_A)^N(\ell_B - \ell_A)]$, and by concavity $F$ exceeds $F(0)$ somewhere iff
$F'_+(0)>0$. If $F'_+(0)>0$ and $F'_-(1)<0$, a maximiser of the concave $F$ cannot be at an
endpoint and is strictly better than both endpoints. Conversely, if $F(\lambda) >
\max(F(0),F(1))$ at an interior $\lambda$, concavity gives $F'_+(0) \ge (F(\lambda)-F(0))/\lambda > 0$, and
symmetrically $F'_-(1)<0$.
(c)~With $X = (1-p_A)^N$ and $Y = (1-p_B)^N$, Cauchy--Schwarz gives $1-F(\tfrac12) = \E\sqrt{XY} \le
\sqrt{\E X\,\E Y} = \E X = 1-F(0)$. Equality iff $X \propto Y$ a.s., i.e.\ $X=Y$, i.e.\ $p_A=p_B$ since
$x\mapsto(1-x)^N$ is injective for $N>0$.
(d)~$\varphi(x,y) = g(x)g(y)$ with $g(p) = (1-p)^{N/2}$ is a product of two non-negative functions
monotone in the same direction, hence supermodular. Gaussian copulas increase in $\rho$ in the
supermodular order \citep{muller2000supermodular}, so $1-F(\tfrac12) = \E\varphi(p_A,p_B)$ is
non-decreasing in $\rho$ while $F(0)$ does not depend on $\rho$.
\end{proof}

With identical marginals the gain decreases to zero as $\rho\to1$; with unequal marginals it is only
non-increasing. We make no analytic claim for imperfect verifiers; the planner evaluates
\cref{eq:law} numerically over mixing weights.

\subsection{Proposition~\ref{prop:reverify} (re-verification)}

\begin{proof}
(a)~For $q\ne\tfrac12$, $B_k(q)\to\ind\{q>\tfrac12\}$ by the law of large numbers; for odd $k$,
$B_k(\tfrac12) = \tfrac12$ exactly by symmetry. Bounded convergence gives the limits.
(b)~$B_k(0)=0$ and $B_k(1)=1$ for all $k$.
(c)~Given the propensities, majority verdicts satisfy \cref{ass:ci} with $a = pT_k$ and
$s = pT_k + (1-p)F_k > 0$, so \cref{prop:law} gives the ceiling
$pT_k/(pT_k + (1-p)F_k) = 1/(1 + \frac{1-p}{p}\cdot\frac{F_k}{T_k})$, strictly decreasing in $F_k/T_k$ for
$0<p<1$. Hence the ceiling with $k$ calls exceeds the ceiling with $j$ calls iff
$F_k/T_k < F_j/T_j$; with $j=1$ this is the negation of \cref{eq:hurt}. Since $F_k/T_k \to
F_\infty/T_\infty$ by (a), the same comparison holds for $k=\infty$. At $p\in\{0,1\}$ the ceiling is
$0$ or $1$ for every $k$. For a joint limit in which $k$ and $N$ grow together while $T_k, F_k \to 0$,
one additionally needs $N(pT_k + (1-p)F_k)\to\infty$ so that abstention vanishes.
\end{proof}

\paragraph{Example and counterexample.} Let $D_w = 0.4\,\delta_{0.05} + 0.6\,\delta_{0.75}$, so 60\% of
wrong answers fool most calls: $F_1 = 0.47$ and $F_\infty = 0.6$. If $D_c = 0.9\,\delta_1 + 0.1\,\delta_0$,
then $T_1 = T_\infty = 0.9$ and the ceiling at $p=0.4$ falls from 0.561 to 0.500. If instead
$D_c = \delta_{0.6}$, then $T_1 = 0.6$ and $T_\infty = 1$, so $T_\infty/T_1 \approx 1.67 > F_\infty/F_1 \approx 1.28$,
and the ceiling rises from 0.460 to 0.526 even though $F_\infty > F_1$.

\subsection{Proposition~\ref{prop:argmax} (argmax selection)}

\begin{proof}
(a)~The maximum score is a.s.\ unique because $H$ is atomless. By exchangeability it belongs to
candidate 1 with probability $1/N$, so $P_N = N\,\Prob(C_1=1, Y_1 = \max_i Y_i) = N\int p\,g_c(y)
H(y)^{N-1}dy$.
(b)~The top occupied bin is $b$ with probability $R_b^N - R_{b-1}^N$ ($R_{-1} = 0$). Conditional on
lying in bin $b$, a candidate is correct with probability $r_c[b]/r[b]$, independently of the
others. Given that $b$ is the top occupied bin, exchangeability and the uniform tie-break give
success probability $r_c[b]/r[b]$. Empty bins contribute zero.
(c)~\emph{Binned:} $R_b < 1$ for $b < b^\ast$ and $R_b = 1$ for $b \ge b^\ast$, so $R_b^N \to
\ind\{b\ge b^\ast\}$ and only the $b^\ast$ term survives. \emph{Continuous:} let $M_N = \max_i Y_i$.
Conditional on $M_N = y$, the maximiser is correct with probability $\eta(y)$, so $P_N = \E[\eta(M_N)]$.
Since $H(y) < 1$ for $y$ below the upper endpoint $y^\ast$, $\Prob(M_N\le y) = H(y)^N \to 0$, so
$M_N \to y^\ast$ in probability. As $\eta\in[0,1]$ converges to $L$ at $y^\ast$, bounded convergence
gives $P_N\to L$. Without the existence of this limit no ceiling of this form follows.
\end{proof}

\subsection{Proposition~\ref{prop:regret} (planner regret)}

\begin{proof}
(a)~Since $\hat c$ maximises $\hat S$, $\hat S(c^\star)\le\hat S(\hat c)$. Hence
$S(c^\star)-S(\hat c) = [\hat S(c^\star)-e(c^\star)]-[\hat S(\hat c)-e(\hat c)] \le e(\hat c)-e(c^\star)$,
which is at most $\max e-\min e\le 2\max|e|$.
(b)~If $\hat c\ne c^\star$, then $\mathrm{Reg}\ge S(c^\star)-\max_{c\ne c^\star}S(c)>\operatorname{osc}(e)$,
contradicting~(a).
(c)~Write $e=b+d$ with $d=S_{\mathrm{cal}}-S$. Oscillation is subadditive, so
$\operatorname{osc}(e)\le\operatorname{osc}(b)+\operatorname{osc}(d)\le\operatorname{osc}(b)+2\max_c|d(c)|$.
For a fixed $c$, $d(c)$ is a difference of two independent means of per-task success values in
$[0,1]$ with equal expectations, so Hoeffding's inequality gives
$\Prob(|d(c)|\ge u)\le2\exp\{-2u^2/(T_{\mathrm{cal}}^{-1}+T_{\mathrm{ev}}^{-1})\}$. A union bound over
$\mathcal C$ and~(a) complete the proof. The bound holds for any estimator: $b$ may depend on the
pilot arbitrarily, because $d$ does not involve the estimator.
\end{proof}

\section{Estimation details}
\label{app:estimation}

\paragraph{Difficulty.} Beta-binomial (optionally zero-inflated) hyper-parameters are fitted by
maximum likelihood on the pilot counts of the calibration tasks. Per-task posteriors are
Beta$(\alpha + k_t, \beta + n - k_t)$, or the corresponding mixture when zero-inflated. The NPMLE
uses EM on a fixed grid of 200 support points that is dense near 0 and 1 and includes the endpoints.

\paragraph{Copula.} Pairwise correlations are fitted by maximising the bivariate beta-binomial
likelihood under a Gaussian copula, using $48\times48$ Gauss--Hermite quadrature. This is accurate
for pilot-scale $n$ (up to a few hundred). The matrix is projected to the nearest correlation
matrix, and $2^{12}$--$2^{14}$ Monte Carlo draws of $(p_1,\dots,p_M)$ represent the prior.

\paragraph{Verifier.} The pooled model estimates the TPR from the accepted fraction of correct
pilot samples, and a beta prior over per-task FPRs by beta-binomial maximum likelihood on the
accepted counts of wrong pilot samples. The task-varying model fits beta-binomial marginals for $p$,
TPR and FPR and a $3\times3$ Gaussian-copula correlation from per-task pilot counts. Each task's TPR is
observed through its correct samples and its FPR through its wrong ones. For LLM judges, a
candidate's propensity is the fraction of the eight judges that accept it. The concentrations
$\kappa_c$ and $\kappa_w$ are fitted by beta-binomial likelihood on these acceptance counts
(typical values: $\kappa_c\approx2.5$, $\kappa_w\approx14$ on MATH-500; $\kappa_c\approx4$,
$\kappa_w\approx60$--$270$ on GPQA). The simulated verifier of \cref{tab:regret-sampling} uses
$\tau=0.9$, per-task FPR $\sim\mathrm{Beta}$ with mean 0.1 and concentration 2, and $\kappa=4$.

\paragraph{Automatic verifier-model choice.} Both verifier models are fitted on the first half of
each calibration task's pilot samples. Each predicts verified success at the full pilot size. We
keep the model whose prediction is closer to the model-free success of uniform-among-accepted
selection on the full pilot.

\paragraph{Argmax.} Verifier probabilities are binned into 20 equal-width bins on $[0,1]$. The
issue-level model is $y_{ti} = \mu_{C_{ti}} + u_t + e_{ti}$ on logit scores, with a random intercept
$u_t\sim\mathcal N(0,s_u^2)$ fitted by EM. Class-conditional residual distributions are kept
empirical. Predictions combine the posterior of $p_t$ with the shifted class-conditional bin
probabilities.

\section{Additional results}
\label{app:extra}

\begin{table}[h]
\centering
\caption{\textbf{Choosing the verifier model automatically} (extrapolation MAE in points, mean over
pilot sizes). Paired tests: the automatic choice is significantly better than the worse fixed
model on every dataset ($p\le0.03$), ties the better one on HumanEval ($p=0.52$), and is behind it on
CodeContests and SWE-bench (0.28 and 0.71 points, $p<0.01$).}
\label{tab:auto}
\small
\begin{tabular}{@{}lcccc@{}}
\toprule
Dataset & Pooled verifier & Task-varying & Automatic & Picks task-varying \\
\midrule
CodeContests (3 models) & \textbf{2.13} & 2.64 & 2.40 & 43\% \\
HumanEval (17 models) & 1.85 & \textbf{1.69} & 1.73 & 56\% \\
SWE-Bench Verified (3 thresholds) & 5.35 & \textbf{2.74} & 3.45 & 59\% \\
\bottomrule
\end{tabular}
\end{table}

\begin{table}[h]
\centering
\caption{\textbf{Agent pool: extrapolation MAE} (points; random splits, 10 seeds). Horizon = all
held-out systems.}
\label{tab:pool}
\small
\begin{tabular}{@{}lcccccc@{}}
\toprule
Pool, pilot & Horizon & Beta prior & NPMLE prior & Power law & Plug-in & Independence \\
\midrule
All 175 systems, 16 & 159 & 3.23 & 2.58 & \textbf{1.71} & 4.93 & 8.40 \\
All 175 systems, 32 & 143 & 2.84 & \textbf{1.27} & 1.58 & 2.56 & 7.51 \\
90 single-attempt systems, 32 & 58 & 2.39 & 1.14 & \textbf{0.73} & 2.66 & 7.87 \\
41 mini-SWE-agent runs, 16 & 25 & 3.39 & 1.65 & \textbf{1.48} & 2.68 & 10.95 \\
\bottomrule
\end{tabular}
\end{table}

\begin{table}[h]
\centering
\caption{\textbf{Argmax selection on SWE-Bench Verified} (6-rollout pilot, 10 draws).}
\label{tab:argmax}
\small
\begin{tabular}{@{}lcc@{}}
\toprule
Method & MAE (points) & Error at $N=16$ \\
\midrule
Power law & \textbf{1.6} & $+2.3$ \\
Plug-in per task & 2.1 & $-2.9$ \\
Ours, issue-level score shift & 3.3 & $+2.4$ \\
Ours, pooled score distributions & 4.5 & $+5.3$ \\
Ours, issue shift with Gaussian tails (ablation) & 3.8 & $-7.4$ \\
Ours, joint Dirichlet over bins (ablation) & 20.3 & $+22.6$ \\
\bottomrule
\end{tabular}
\end{table}

\label{app:figures}

\begin{figure}[h]
  \centering
  \includegraphics[width=\linewidth]{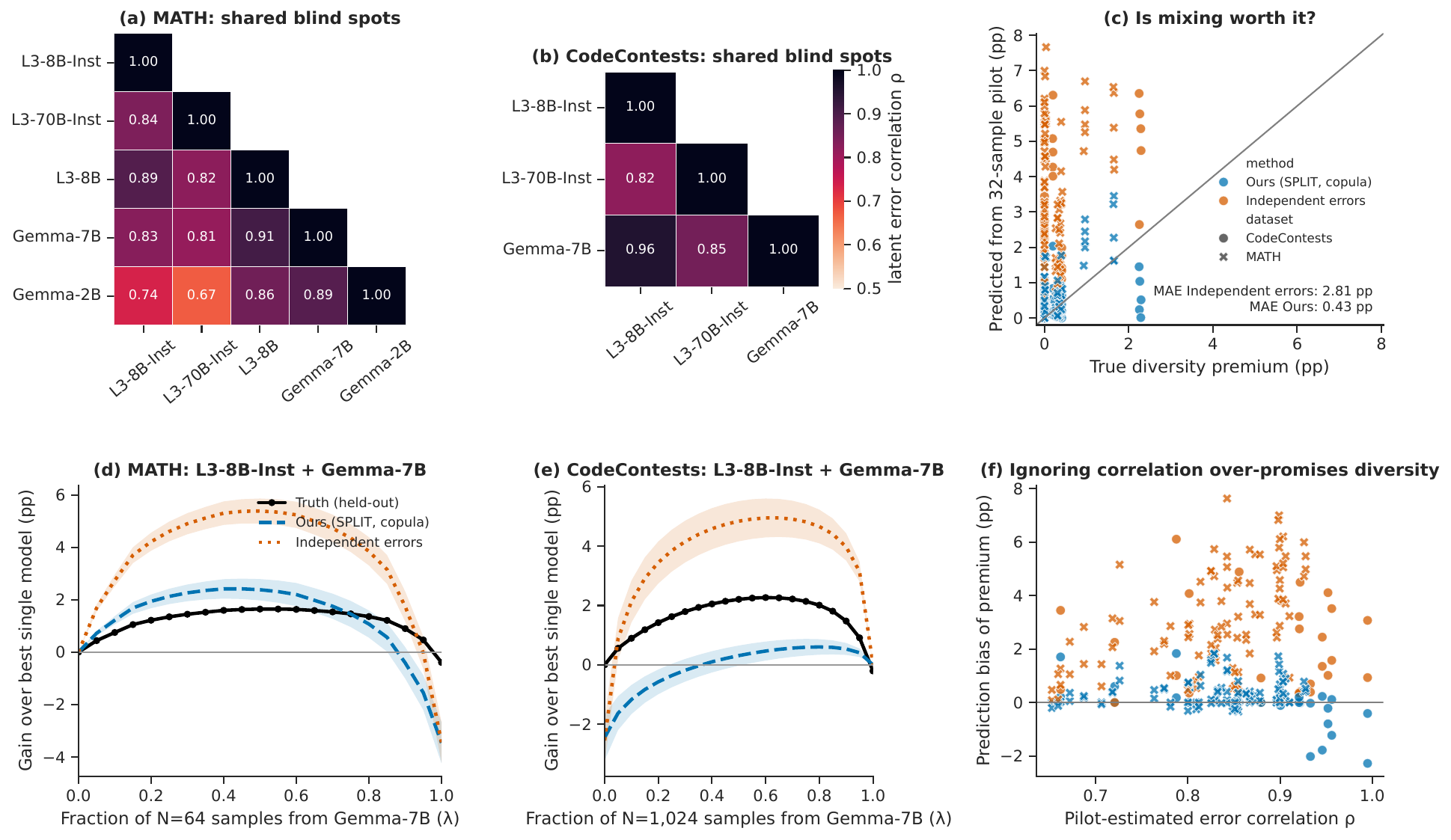}
  \caption{\textbf{Shared blind spots.} (a,~b)~Pilot-estimated copula correlations between generator
  pairs on MATH and CodeContests. (c)~Predicted versus real gain from the best two-model mix: the
  SPLIT with the copula (ours) against the independence assumption. (d,~e)~Mixing curves. (f)~The
  independence over-promise grows with correlation.}
  \label{fig:diversity}
\end{figure}

\begin{figure}[h]
  \centering
  \includegraphics[width=\linewidth]{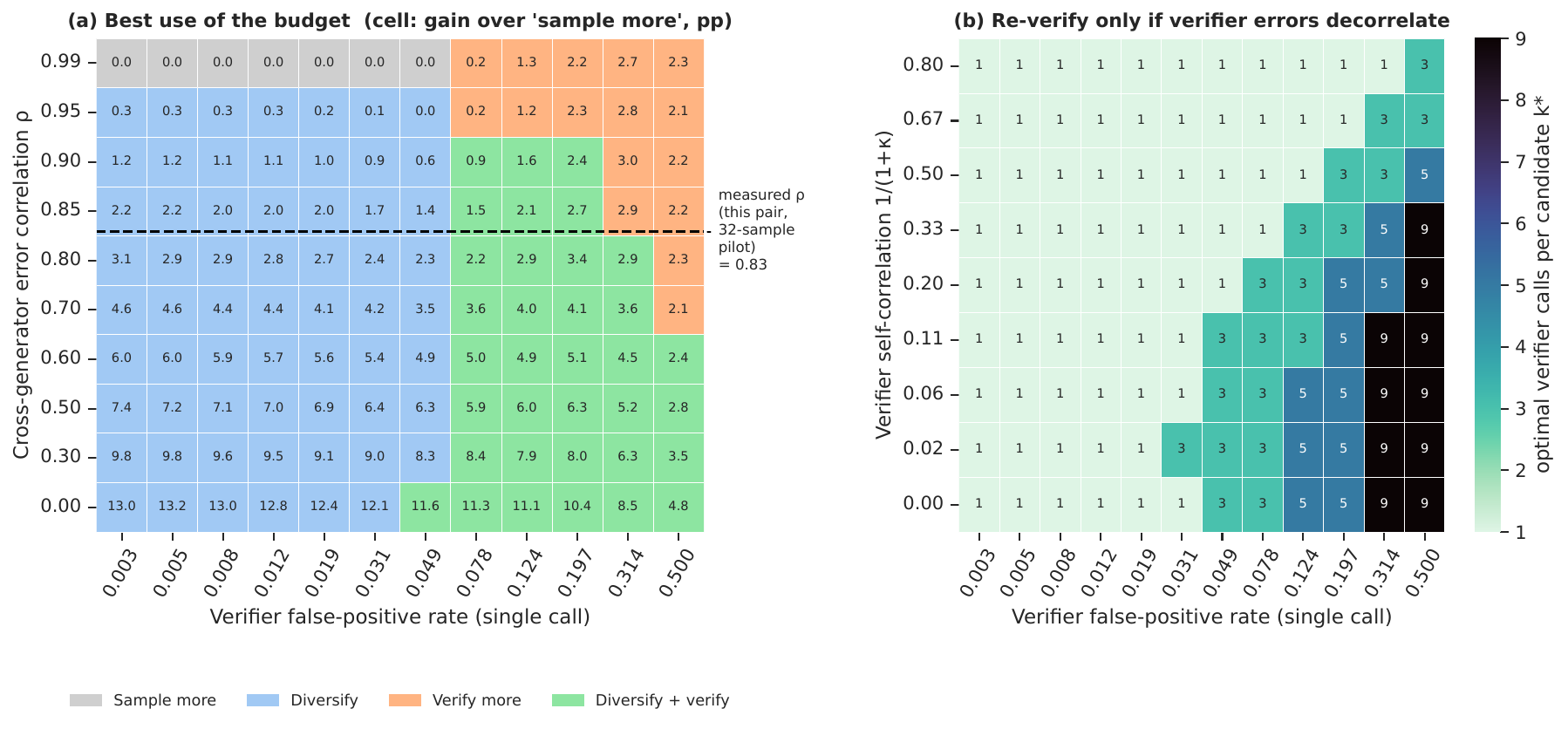}
  \caption{\textbf{Which lever to pull} (theory). (a)~Best use of a fixed budget as a function of
  verifier FPR and cross-generator correlation: sample more, diversify, verify more, or both.
  (b)~The optimal number of verifier calls stays at one when repeated calls are highly correlated.}
  \label{fig:phase}
\end{figure}

\begin{figure}[h]
  \centering
  \includegraphics[width=\linewidth]{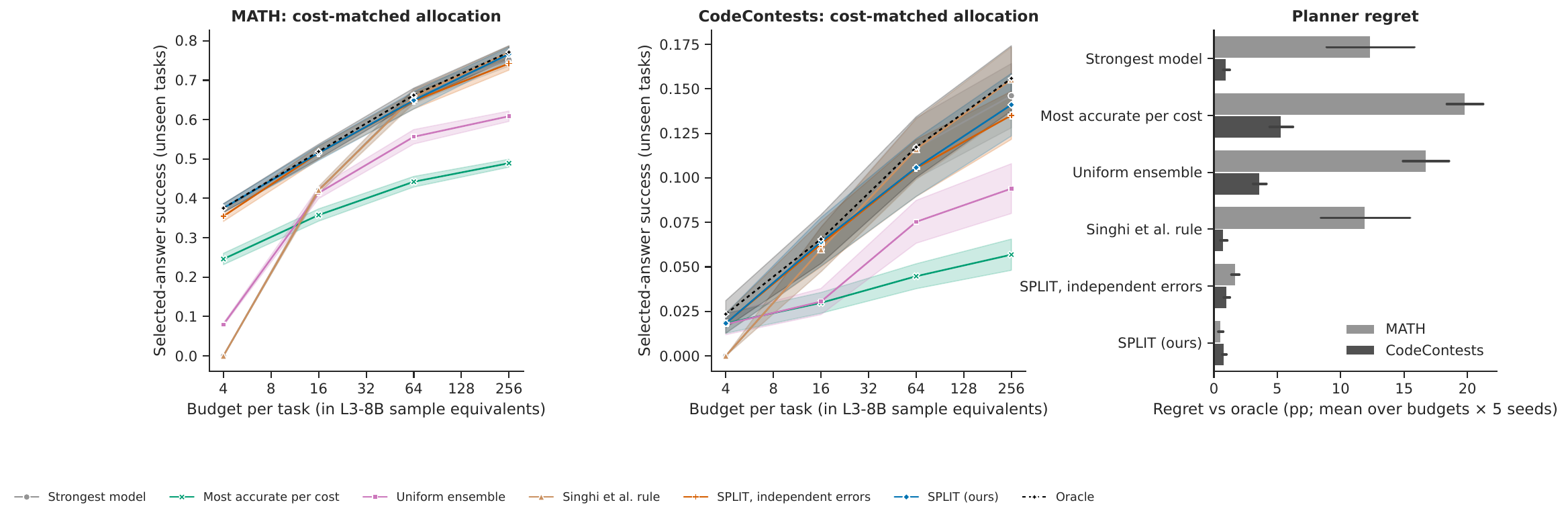}
  \caption{\textbf{Allocation with a simulated verifier} (MATH and CodeContests, 5 splits).
  Success of the chosen configuration on unseen tasks by budget, and regret relative to the oracle.}
  \label{fig:alloc-sim}
\end{figure}

\begin{figure}[h]
  \centering
  \includegraphics[width=0.78\linewidth]{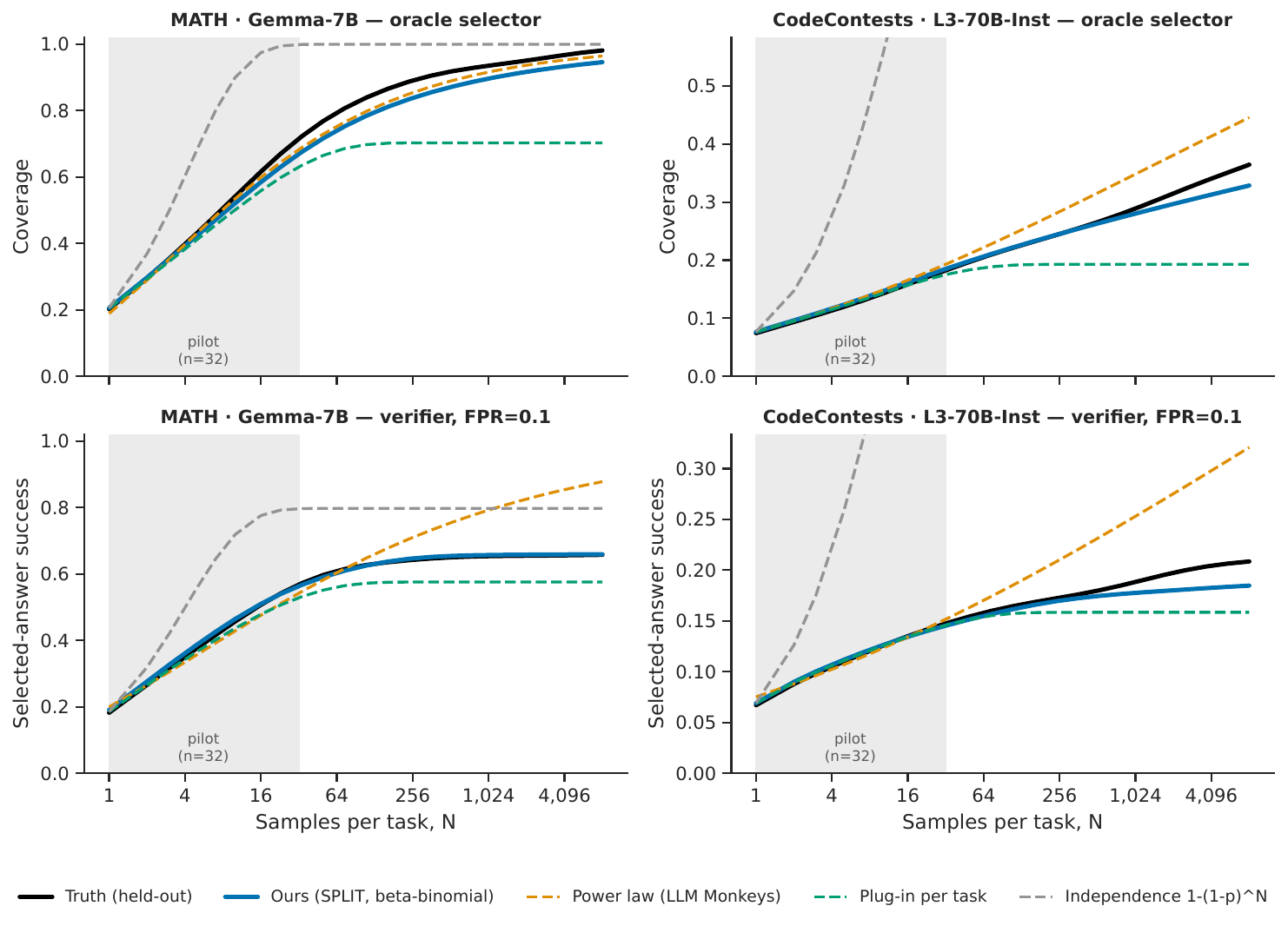}
  \includegraphics[width=0.78\linewidth]{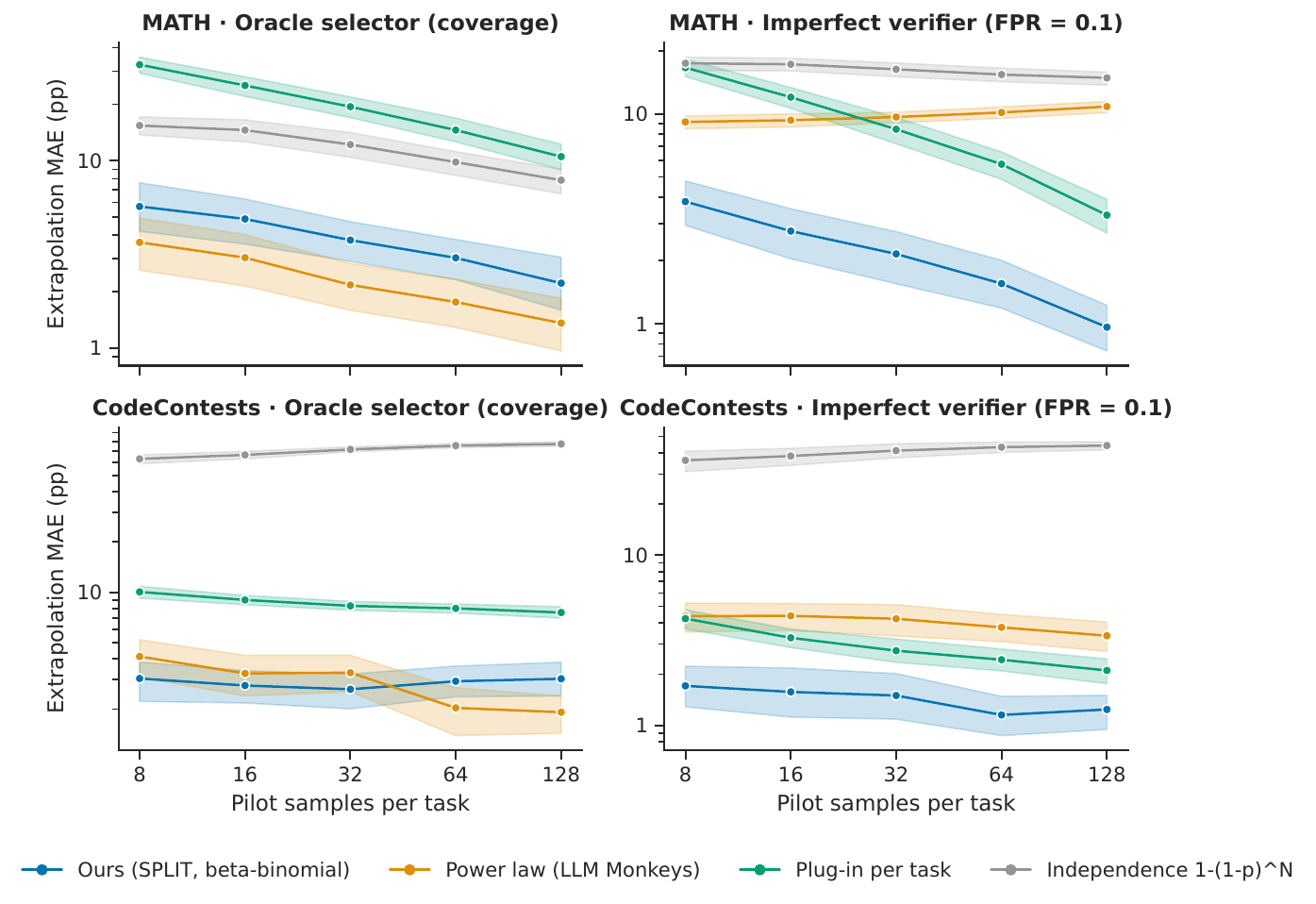}
  \caption{\textbf{Pilot extrapolation with a simulated verifier} (FPR 0.1). Top: predicted curves
  from a 32-sample pilot. Bottom: error against pilot size for coverage and selection.}
  \label{fig:pilot-sim}
\end{figure}

\begin{figure}[h]
  \centering
  \includegraphics[width=\linewidth]{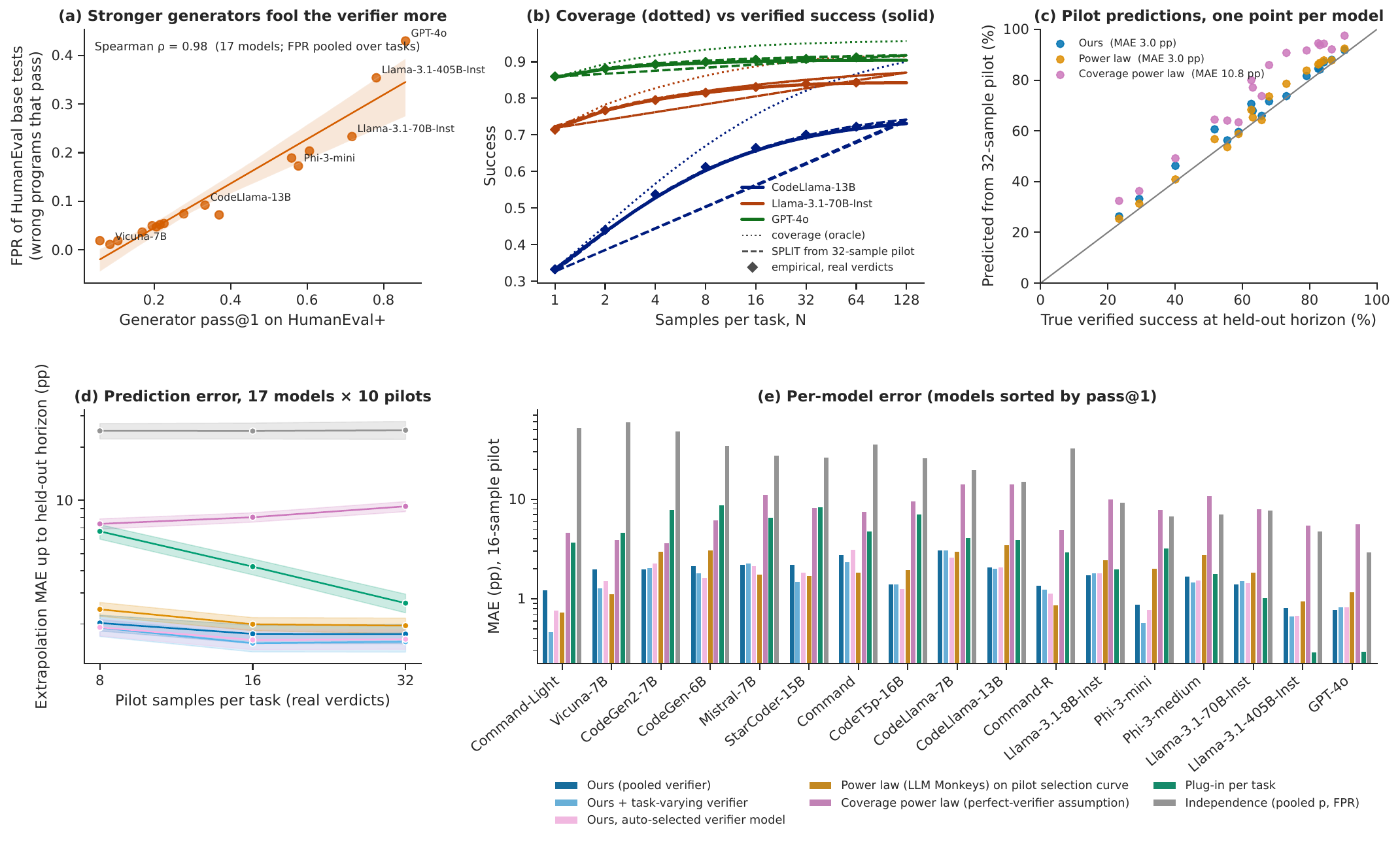}
  \caption{\textbf{HumanEval base tests as a real verifier, HumanEval+ as ground truth, 17
  generators.} (a)~FPR, pooled over tasks, against pass@1 (Spearman 0.98; 0.97 with per-task FPRs as
  in \cref{sec:res-mech}). (b)~Coverage and verified success for three generators.
  (c)~Predicted versus true success at the held-out horizon. (d)~Error against pilot size. (e)~Error
  per generator.}
  \label{fig:humaneval}
\end{figure}

\begin{figure}[h]
  \centering
  \includegraphics[width=\linewidth]{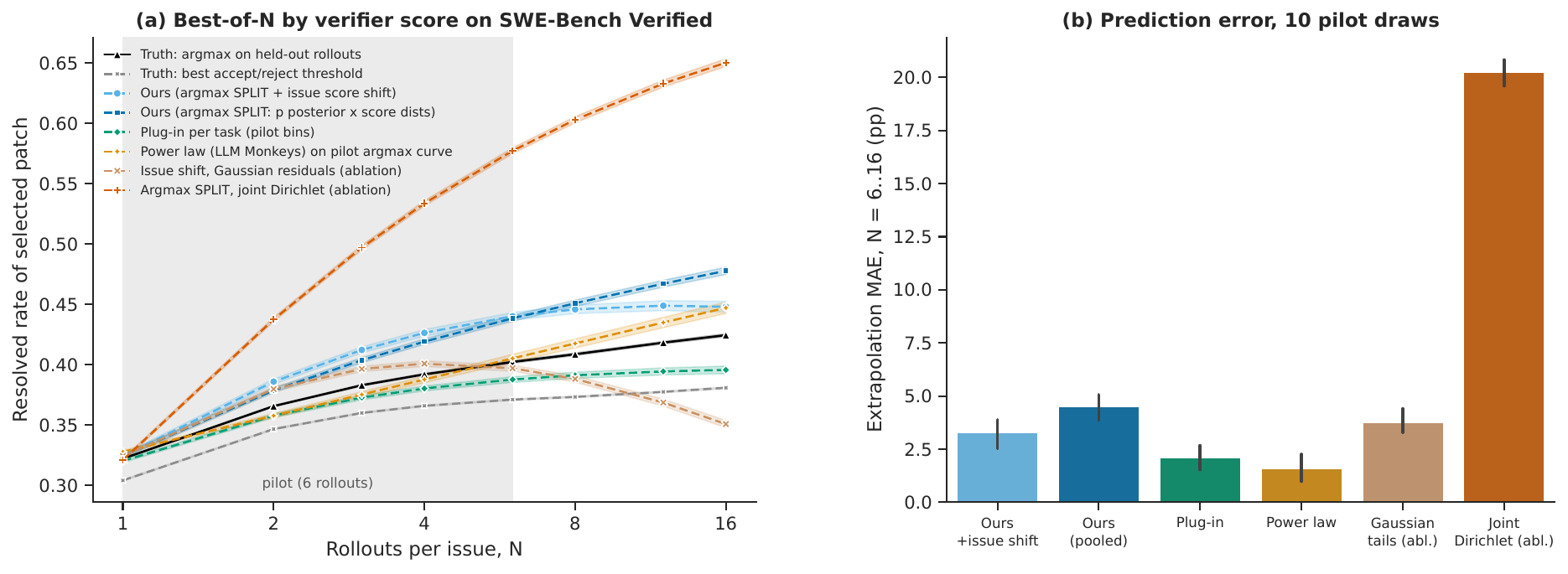}
  \caption{\textbf{Argmax (best-of-$N$) selection on SWE-Bench Verified.} (a)~Truth and predictions;
  the best fixed accept/reject threshold (grey; 0.3, 0.5 or 0.7, scored as a failure when no rollout
  is accepted) is below argmax at every $N$, by 1.8 points at $N=1$ and 4.4 at $N=16$.
  (b)~Extrapolation error.}
  \label{fig:argmax}
\end{figure}

\begin{table}[h]
\centering
\caption{\textbf{Planner-regret diagnostics} (\cref{prop:regret}; percentage points, mean over budgets and
splits). Pairwise error is $e(\hat c)-e(c^\star)$, which bounds regret. Model spread is
$\operatorname{osc}(b)$, the variation across configurations of SPLIT's error on the calibration
tasks. Task gap is $\max_c|S_{\mathrm{cal}}(c)-S(c)|$. Hoeffding is the finite-sample term of
\cref{prop:regret}(c) at $\delta=0.05$.}
\label{tab:regret-diag}
\small
\setlength{\tabcolsep}{4pt}
\begin{tabular}{@{}lrrcccccc@{}}
\toprule
Setting & $|\mathcal C|$ & $T_{\mathrm{cal}}$ & Regret & Pairwise error & $\max|e|$ & Model spread & Task gap & Hoeffding \\
\midrule
MATH, simulated verifier & 140 & 64 & 0.54 & 0.96 & 11.12 & 7.20 & 10.40 & 73.4 \\
CodeContests, simulated verifier & 48 & 70 & 0.82 & 2.81 & 7.59 & 3.69 & 5.79 & 65.7 \\
CodeContests, real tests & 13 & 70 & 0.62 & 1.48 & 6.03 & 2.63 & 5.50 & 59.8 \\
\bottomrule
\end{tabular}
\end{table}

\clearpage
\section{Significance tests}
\label{app:significance}

\begin{table}[h]
\centering
\caption{\textbf{Paired tests for headline comparisons.} Mean difference (ours minus comparator;
negative favours ours) with bootstrap 95\% interval and Wilcoxon signed-rank $p$-value. Units are
shared (model, pilot, seed) for MAE and (dataset, seed, budget) for regret.}
\label{tab:sig}
\small
\setlength{\tabcolsep}{4pt}
\begin{tabular}{@{}lcc@{}}
\toprule
Comparison & Mean diff.\ [95\% CI] & $p$ \\
\midrule
Selection MAE, sim.\ verifier: SPLIT vs power law & $-5.71$ [$-6.17$, $-5.22$] & $<10^{-3}$ \\
CodeContests MAE: SPLIT vs power law & $-0.65$ [$-1.00$, $-0.32$] & 0.002 \\
HumanEval MAE: SPLIT, task-varying vs power law & $-0.44$ [$-0.55$, $-0.33$] & $<10^{-3}$ \\
SWE-bench MAE: SPLIT, task-varying vs power law & $-3.97$ [$-4.34$, $-3.59$] & $<10^{-3}$ \\
SWE-bench argmax MAE: SPLIT, issue shift vs power law & $+1.71$ [$+1.00$, $+2.35$] & 0.006 \\
Agent-pool coverage MAE: SPLIT, NPMLE vs power law & $+0.30$ [$-0.11$, $+0.72$] & 0.57 \\
Agent-pool $|$error$|$ at max $N$: SPLIT, NPMLE vs power law & $+0.99$ [$+0.46$, $+1.56$] & 0.004 \\
\midrule
Regret, sim.: SPLIT vs independent errors & $-0.67$ [$-1.10$, $-0.30$] & 0.003 \\
Regret, sim.: SPLIT vs strongest model & $-5.99$ [$-10.03$, $-2.47$] & 0.009 \\
Regret, sim.: SPLIT vs \citet{singhi2025solve} & $-5.69$ [$-9.87$, $-2.08$] & 0.13 \\
Regret, sim., compute-matched: SPLIT vs \citet{singhi2025solve} & $-5.78$ [$-9.98$, $-2.16$] & 0.09 \\
Regret, real tests: SPLIT vs independent errors & $-0.84$ [$-1.22$, $-0.50$] & $<10^{-3}$ \\
Regret, real tests: SPLIT vs strongest model & $-0.32$ [$-0.92$, $+0.23$] & 0.43 \\
Regret, judges: empirical task population vs \citet{singhi2025solve} & $-0.58$ [$-0.76$, $-0.42$] & $<10^{-3}$ \\
Regret, judges: empirical task population vs judge-level IRT & $-0.00$ [$-0.03$, $+0.02$] & 0.93 \\
Regret, SWE-bench: SPLIT, task-varying vs fixed threshold 0.5 & $-2.29$ [$-3.12$, $-1.52$] & $<10^{-3}$ \\
Regret, SWE-bench: SPLIT, task-varying vs pooled verifier & $-2.03$ [$-3.37$, $-0.88$] & 0.005 \\
\midrule
Regret, real tests, $\gamma=0.7$: SPLIT vs strongest model & $-0.32$ [$-0.90$, $+0.21$] & 0.55 \\
Regret, real tests, $\gamma=0.7$: SPLIT vs most accurate per cost & $-4.13$ [$-5.28$, $-2.92$] & $<10^{-3}$ \\
Regret, real tests, $\gamma=0.5$: SPLIT vs strongest model & $+0.45$ [$+0.22$, $+0.72$] & $<10^{-3}$ \\
Regret, real tests, $\gamma=0$: SPLIT vs strongest model & $+0.50$ [$+0.26$, $+0.80$] & 0.001 \\
Regret, sim., $\gamma=0.7$: SPLIT vs \citet{singhi2025solve} & $-4.71$ [$-9.01$, $-1.02$] & 0.59 \\
Regret, online: re-planning (pooled) vs static plan & $-0.56$ [$-0.81$, $-0.31$] & $<10^{-3}$ \\
Regret, online: re-planning (pooled) vs static oracle & $+0.06$ [$-0.34$, $+0.49$] & 0.48 \\
Regret, online: re-planning vs early stopping (both pooled) & $-0.19$ [$-0.54$, $+0.15$] & 0.28 \\
\bottomrule
\end{tabular}
\end{table}

\section{Data, compute and reproducibility}
\label{app:repro}

\paragraph{Data sources.} MATH and CodeContests samples: Large Language Monkeys release
\citep{brown2024monkeys}. HumanEval and MBPP samples with EvalPlus verdicts: release accompanying
\citet{stroebl2024limits}. MATH-500 and GPQA candidates with LLM-judge verdicts, reward-model
scores and Weaver scores: Weaver release \citep{saadfalcon2025weaver}. SWE-Bench Verified rollouts
with verifier probabilities: R2E-Gym release \citep{jain2025r2egym}. Per-instance results of 175
systems: SWE-bench experiments repository \citep{swebenchexperiments}. We use these releases under
their respective licences. We do not redistribute model outputs. The code release includes only
derived per-sample correctness labels and verifier verdicts, so that the main experiments run
without the multi-gigabyte downloads.

\paragraph{Compute.} All experiments run on a single 8-core laptop CPU. Executing the CodeContests
programs takes about 1.5 hours; the remaining experiments take roughly 5--8 hours in total. No GPU
or model API is used.

\paragraph{Reproducibility.} The code will be released publicly. It contains the estimators, the
planner, the online policies, scripts that regenerate every number and figure in this paper from
the public releases, 35 unit tests, and 26 numerical checks of the propositions. Random seeds are
fixed. Token counts for compute-matched and price costs are measured from the released samples.

\paragraph{Long same-agent rollouts.} The code release also contains a pipeline for the experiment
that public data cannot support: collect $R$ rollouts of one fixed agent per SWE-Bench Verified
issue with a standard open-source scaffold, score them with the official evaluation harness and an
execution-free LLM verifier ($P(\text{yes})$ from the first-token log-probabilities), and re-run the
threshold and argmax analyses with pilots of 4--16 rollouts and horizons up to $R-16$. At $R=64$
this is 32{,}000 agent runs. A self-test runs the analysis path on resampled public rollouts.

\end{document}